\documentclass[12pt,letterpaper,twoside]{report}
\usepackage[left=1.25in,right=1.25in,top=1.35in,bottom=1.35in,bindingoffset=0.35in]{geometry}

\usepackage[utf8]{inputenc}
\usepackage{verbatim} 
\usepackage{amsmath}
\usepackage{amsfonts}
\usepackage{amssymb}
\usepackage{numprint}
\npdecimalsign{.}
\nprounddigits{2}

\usepackage{cite} 

\ifx\pdftexversion\undefined
\usepackage[dvips]{graphicx}
\else
\usepackage[pdftex]{graphicx}
\fi
\usepackage{titlesec}
\usepackage{epstopdf} 
\usepackage{color}
\usepackage{booktabs}
\usepackage{float}
\usepackage{url}

\usepackage[stable]{footmisc} 

\usepackage{subfigure}
\usepackage{placeins} 
\usepackage{listings}
\usepackage{algorithm2e}
\usepackage{todonotes}
\usepackage{acronym}
\usepackage{hyperref} 

\usepackage[english]{babel}
\usepackage[T1]{fontenc}

\usepackage{lmodern}	
\usepackage{amsmath}	
\usepackage{amsthm}
\usepackage{amsfonts}
\usepackage[toc]{appendix} 

\usepackage{hhline}     
\usepackage{tablefootnote}

\begin{document}
\begin{titlepage}

\centering

\includegraphics[width=100mm]{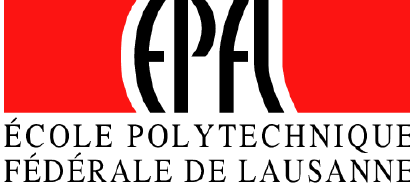}\\

\vskip 4.5cm

\hrule
\vskip 0.5cm
{\Huge \sffamily Deep Learning applied to Image and Text matching\\}
\vskip 0.5cm
\hrule
\vskip 0.5cm
{\large \sffamily MASTER THESIS IN COMPUTER SCIENCE}
\vskip 3.15cm

\begin{tabular}{lr}
\begin{minipage}[t]{0.5\linewidth}
\Large \sffamily
\textbf{Author:}\\[3pt]
Afroze Ibrahim Baqapuri\\afroze.baqapuri@epfl.ch\\[3pt]
\end{minipage}
&
\begin{minipage}[t]{0.5\linewidth}
\Large \sffamily
\begin{flushright}
\textbf{Supervisors:}\\[3pt]
Dr. Fran\c{c}ois Fleuret\\francois.fleuret@idiap.ch\\[3pt]
Dr. Eric Cosatto\\cosatto@nec-labs.com\\[3pt]
\end{flushright}
\end{minipage}
\end{tabular}

\vfill \Large \sffamily \today

\end{titlepage}

\onecolumn\cleardoublepage
\tableofcontents\cleardoublepage


\cleardoublepage
\chapter*{Abstract}
The ability to describe images with natural language sentences is the hallmark for image and language understanding. Such a system has wide ranging applications such as annotating images and using natural sentences to search for images. In this project we focus on the task of bidirectional image retrieval: such a system is capable of retrieving an image based on a sentence (image search) and retrieve sentence based on an image query (image annotation). We present a system based on a global ranking objective function which uses a combination of convolutional neural networks (CNN) and multi layer perceptrons (MLP). It takes a pair of image and sentence and processes them in different channels, finally embedding it into a common multimodal vector space. These embeddings encode abstract semantic information about the two inputs and can be compared using traditional information retrieval approaches. For each such pair, the model returns a score which is interpretted as a similarity metric. If this score is high, the image and sentence are likely to convey similar meaning, and if the score is low then they are likely not to.

The visual input is modeled via deep convolutional neural network. On the other hand we explore three models for the textual module. The first one is bag of words with an MLP. The second one uses n-grams (bigram, trigrams, and a combination of trigram \& skip-grams) with an MLP. The third is more specialized deep network specific for modeling variable length sequences (SSE). We report comparable performance to recent work in the field, even though our overall model is simpler. We also show that the training time choice of how we can generate our negative samples has a significant impact on performance, and can be used to specialize the bi-directional system in one particular task.

\cleardoublepage
\chapter*{Acknowledgments}
I would like to thank NEC and EPFL for giving me the opportunity to work on such an interesting project. Special thanks go to Bing Bai, Iain Melvin, Eric Cosatto, Igor Durdanovic, and Martin Renqiang Min for the many fruitful discussions.

\cleardoublepage
\chapter{Introduction}\label{ch:introduction}

The ability to describe images with natural language sentences is the hallmark for image and language understanding. It has significant applications such as searching images with natural sentences, and automatic captioning of images. Currently, commercially available image searching systems search the text adjacent to an image, instead of searching the content of the image itself. This is a severe bottleneck on the performance of these systems. Advances in this field will have far-reaching effects since images are ubiquitous on the Internet and we are always interacting with them.

\bigskip
Humans can describe an image with relative ease. However, for computers this is not a trivial task. The difficulty arises mainly because the two input modalities have very different statistical properties, and cannot be directly compared. For example, images have spacial dependency, while sentences have temporal dependency (sequence). But, even for humans, the task is of highly subjective nature, and various forms of descriptions exist. Different descriptions could focus on different aspects, or different objects in the image, they could also have different level of detail in describing the content, and they could be abstract like describing the mood or concrete like describing objects. All of these could be correct simultaneously correct, yet very different. Modeling this variance is very important for machine learning systems, and it is achieved in high-quality data sets by having multiple humans describe the same image.

\bigskip
The recent work in this field has focus on two approaches: Multimodal retrieval and sentence generation. Multimodal retrieval deals with the tasks of retrieving sentences from image queries (or \textbf{image captioning}) and retrieving images from sentence queries (or \textbf{image search}). On the other hand, image based sentence generation systems focus on creating natural and fluent sentences directly from the image, which describe the contents of the image.

\bigskip
In our work, we focus on the former approach: multimodal retrieval. We design our system taking motivation from traditional retrieval systems, in particular the supervised semantic indexing (SSI) \cite{text-SSI} system at NEC used for document retrieval. We expand on the system by using separate, specialized textual and visual models. We use these models to extract semantic information from the input sentence and image, and then transform it into a common vector space, where we can easily compare the two modalities using traditional information retrieval approaches. 

\bigskip
The goal of our project is to explore and understand the significance of deep learning techniques in this task. We use deep learning architectures and concepts within the individual text and image modules. For the visual component we experiment using deep convolutional neural networks, which have become synonymous with image recognition \cite{vision-krizhevsky,vision-document-lenet5-lecun}. For the textual component we experiment with different features and network architectures. We try bag of word approach, n-gram models, tf-idf features and convolutional neural network for sequence embedding. In contrast to other previous work, our model is simpler since we don't use more complicated models like recurrent neural networks \cite{model-brnn-feifei} or recursive neural networks \cite{model-SDT_RNN-socher}. However, even with the simpler approach, our model achieves comparable performance to some recent work in the field.

\section{Organization of report}
The rest of the thesis report is organized as follows. In Chapter \ref{ch:lit_review} we present a through overview of available scientific literature in the field. We start with deep learning from a historical perspective and discuss its main ideas and significance, first in the image community and then in text processing. Finally we discuss the recent previous work in the combined field of images and texts as a multimodal input. 

Following this, In Chapter \ref{ch:resources} we briefly describe some of the important resources used in our project. This includes data sets we trained and tested our model on and some publicly available tools we used in our larger model.

In Chapter \ref{ch:methodology} we provide an overview of research methodology and time line of experiments we conducted. This chapter serves as an optional section. So the reader can skip over it without loosing necessary information in understanding our final models and experiments (reference to the chapter will be made whenever we make a point which needs support from there). The material covers our experience with the problem how it influenced the way we approached the task (including the set backs we faced).

Chapter \ref{ch:proposed_model} presents an overview and description of the model we use in our retrieval system including flow charts and mathematical description. It also explains some interesting features of the model, and how does it compare with recent work.

Moving forward, Chapter \ref{ch:experiments} deals with the bulk of the experimentation done on our model, and the results we get. We start off by discussing the preprocessing, then give details on the evaluation metrics used and meta parameters selected. Finally we conduct several experiments and report their results.

In Chapter \ref{ch:results} we sum up our results by comparing them with other recent models. We then discuss the significance of our results and possible future work in this direction.

Finally, in Appendix, we give a brief overview of the software environment we used to train and test our models, and some additional resources we used which did not become part of our final model.

\cleardoublepage
\chapter{Literature Review}\label{ch:lit_review}

In this section we will provide a summary of important research conducted in the field of deep learning during the last decade. The literature review will comprise of three parts:
\begin{enumerate}
\item Deep learning in images and computer vision.
\item Deep learning in text and NLP.
\item Deep learning in image and text multimodal modeling.
\end{enumerate}
The last part is most relevant to our task, but we feel a background discussion is important to understand and appreciate the workings of it.

\section{Deep learning in images and computer vision}

An important aspect of deep learning is to use end-to-end automatically trainable systems which do not rely on human-designed heuristics. Traditional machine learning systems are divided into two modules. First, the $feature extractor$ module transforms the input data into low dimensional vectors which can be easily matched and compared, and which are relatively invariant to distortions. These are then fed into the $classifier$ module, which is general-purpose and trainable. A major problem with this approach is that performance is largely determined by human input, and the feature extractor part is task-specific so it needs to be redone for every little task.

Countering this traditional approach, \cite{vision-document-lenet5-lecun} showed that hand-crafted feature extraction can be advantageously replaced with automatic learning algorithms which operate directly on raw input data. The individual modules can thus be replaced by unified system which optimizes a global performance criterion. 

\bigskip
Computer vision, and especially object recognition, has a long history of using deep multi-layered neural networks. \cite{vision-comparison-lecun} did a comparison of hand-written digit recognition on the MNIST data set, in which they compare the performance of a multi-layered (deep) convolutional neural network (CNN) with traditional  machine learning approaches like linear classifier (logistic regression), principal component analysis, and nearest neighbour classifier. The comparison shows that deep CNNs outperform the traditional algorithms on object recognition tasks, and reach state-of-the-art.

The convolutional neural networks are one of the first used deep learning models, and they are biologically inspired variants of the ANN. \cite{hubel1968receptive} found the existence of receptive fields (small sub-regions in the visual field) in the visual cortex of brain, and CNNs (among other models) try to emulate this behaviour. In other words, they are specialized network architectures specifically designed to recognize two dimensional objects, while being invariant to exact position of the pattern and distortions. In the CNN each unit takes its input from a local receptive field on the layer below forcing it to extract local features. Furthermore units within a plane or $feature map$ are constrained to share a single set of weights, this makes the operations performed by a feature map to be shift invariant \cite{le1990handwritten}. The $weight-sharing$ technique also reduce the number of free parameters, thereby reducing the memory requirements and training complexity of the networks.

Complete CNNs are formed by stacking together multiple convolutional layers (each with $feature map$ planes and local $receptive fields$). Sub-sampling layers are also added improving invariance to shift and distortions. The entire network is trainable with gradient descent  using the back-propagation procedure. \cite{vision-document-lenet5-lecun} popularized the LeNet-5 which is a CNN architecture performing state-of-the-art on MNIST hand writing recognition at the time it was published. 

Even during 1990's it was evident that deeper and larger networks have the tendency to perform better. However, this potential was overshadowed by the outrageous time and memory which took to train larger networks - and was not possible during that time. Moreover, larger networks - being more powerful at modeling - also easily overfit to training data, and there were not efficient techniques to combat this annoyance. Even though near human performance was reached for simple task like hand written recognition, but this was not reciprocated to objects recognition in realistic settings which exhibit considerable variability.

\bigskip
In 2009 \cite{resources-imagenet-feifei} released teh ImageNet data set comprising over 15 million high-resolution images labeled into more than 22,000 categories. It has been well understood that training complicated data with high variability using very few examples leads to severe overfitting in the CNNs, so the release of this large-scale data set was a big step for object recognition problem. By this time computer hardware technology had also progressed enough to train much larger and deeper networks in reasonable amount of time. \cite{vision-krizhevsky} used graphical processor units (GPUs) for a very fast an efficient implementation of their CNN, which has 650,000 neurons and 60 million parameters (in contrast the LeNet-5 had 60,000 free parameters). They entered their neural network in the ILSVRC-2012 competition and achieved a winning top-5 test error rate of 15.3\%, which was considerably better than the state of the art (second-best entry had 26.2\% error rate).

Besides larger data set and larger networks, they also empirically showed the importance of some useful techniques for
avoiding overfitting. The easiest illustrated way is to use data augmentation to increase the size of data set. They take five different patches of the original image (along with their horizontal reflections) thus increasing the training data by a factor of 10 - although the additional images are very close the original one. They also also alter intensities of the colour channels to add further data augmentation. These techniques are useful for adding more shift, inversion, illumination, and colour invariance to our model. Dropout \cite{hinton2012dropout} is another technique for combating overfitting, by reducing complex-co adaptions of neurons. Therefore a single neuron is forced to learn more robust features without relying on other neighbouring neurons. Pretraining the neural network in a greedy layer-by-layer fashion with an unsupervised objective function is another popular technique \cite{hinton2006fast,scholkopfgreedy}. The intuition behind this idea is that unsupervised training will give a good initialization of weights for the neural network based on the actual statistical properties of the data it will be used for (e.g. object images, human speech, etc.) instead of random initializations which often get stuck in poor local minimas. Following this the network can be $fine-tuned$ on a supervised task such as object recognition. 

Mathematically speaking, the CNN transforms the into a low dimensional feature vector representation. In this way a good CNN model can also act as a good feature extractor for images, and the resulting images can be used in more complicated tasks. \cite{resource-overfeat} display this concept for object localization. They train a CNN for classification on the ImageNet data set, and for localization take they replace the final classification layer with a $regression network$. This regression network is simply an MLP with two hidden layers of 4,096 and 1,024 units, connected finally to the output layer of 4 units which predicts the coordinates of the bounding boxes. The final layer is class-specific having 1,000 versions (one for each class) while the rest of the regression network shares weights. During localization only the regression network's weights are updated, and the remaining larger network only acts as a feature extractor. They run the classifier and the regressor simultaneously since they share most of the network, in this way they get a bounding box for each class along with a confidence number based on classification confidence.

\section{Deep learning in text and NLP}

A seminal paper in the domain of statistical learning applied to natural language processing (NLP) was written by \cite{text-neural_language_model-yoshua}. Classical machine learning approaches to NLP calculate n-gram conditional probabilities based simply on the co-occurrence frequencies of words in a document. These models construct tables of of conditional probabilities for a next given a fixed context (of previous $n-1$ words). A fundamental problem with this is the \textit{curse of dimensionality}, meaning that the possible combinations of contexts grows exponentially with $n$, thus making the models quickly intractable. Another problem is that as the size of context increases, the occurrence of sequences gets extremely rare in a document, thus undermining the statistical relevance of their probability distributions. 

The authors of the paper attempt to solve this problem by using a neural network to to learn \textbf{distributed representation} of words. This distributed representation, which is sometimes termed \textbf{word embedding} in modern literature, is feature vector which conveys semantic and syntactic meaning of the word. In other words, all the words in the vocabulary can be transformed into a vector representation of a fixed dimension (usually much smaller than the original size of the vocabulary). Furthermore the joint probability of the entire sequence (of arbitrary length) can be expressed as a function of these feature vectors. In this paper they propose to use feed forward neural networks to compute the probability of the next word in the sequence given the previous $n$ words (in the form of these word vectors) . The advantage of neural networks is that they can be trained in trained to jointly learn the word feature vectors (first layer of the neural network) and the parameters of the probability function (free parameters of the network) using a common global objective function (maximizing log likelihood).

The advantage of using these word vectors is that it would escape the curse of dimensionality. The words would now be represented in a fixed real-valued (comparatively) low-dimensional vector, and similar words would have similar meanings. An example given by the paper is that the sentences "A cat is walking in the bedroom" would have similar representation to "The dog was running in a room", since the model would learn the similarity of the individual words in the sequence (dog, cat), (the,a), (bedroom, room), etc.  It is noteworthy that they report 24\% and 8\% improvements in terms of perplexity over the best n-gram results for the task of predicting the next word in a sequence performed on two large scale data sets. The idea of using neural networks for language modeling in fact dates back to \cite{miikkulainen1991natural}, however the authors of the paper under discussion were the first to propose a large scale statistical model which learns distributed dense representations of words in a sequence and using them to automatically estimate the joint probability function.

\bigskip
Building on to this, \cite{text-unified-collobert} propose a single unified convolutional neural network architecture that performs well for various challenging NLP tasks, such as part-of-speech tagging, chunking, named entity recognition, and semantic role labeling. Traditional approaches analyze these tasks separately, using hand-crafted features specific for each task which makes this approach intractable for complicated tasks. The model they propose -  in contrast - has a deep architecture composing of many layers which can be trained in an end-to-end fashion. The first layer extracts features for each word (word embeddings, or distributed word representations). The second layer extracts features form the sentence treating it as a sequence with structure. Variable length inputs are incorporated by passing a convolutional neural network over the word features and then performing max-pooling over each resulting dimension to give a fix length vector. The following layers are classical NN layers (fully connected).

Since all of these tasks are related , they argue that it would make sense to share some features between these tasks in order to improve the generalization of the network, and so they propose multitask learning approach to \textit{jointly} train the model on all these tasks. They design their network architecture to share the layers closer to the input (these layers would encode word embeddings, which should intuitively be common to all language-related tasks). As we go deeper into the network, the features extracted become more complex and abstract, and so the last layers of the network are task specific. 

They also pretrain their network with unlabeled training data , since it is available in much vast quantities as compared to labeled data. They train a language model with an unsupervised ranking criterion, which would predict if the middle word in the sequence is related to the context or not. For positive examples they took fixed length phrases from wikipedia, and they generated the negative examples by substituting the middle word in a valid phrase by any other random word. They demonstrated that this approach trains a powerful language model by showing that word vectors which are close to one another in the embedding space are also close in semantic meaning. For example "France", "Spain" and "Italy" have close vector representations to one another, as well as "scratched", "smashed" and "ripped".

\bigskip
\cite{resource-word2vec-related} propose yet another unsupervised approach of learning vector representations of word. They propose a \textbf{skip gram} model, which predicts the surrounding words in the window given the center word. This is directly opposite of earlier approaches which predict the centre word giver the context window. Given a sequence of training words, the objective criterion is to maximize the average log probability of the words in surrounding, conditional on the centre word. One interesting feature about their model is that it preserves linear regularities among the learned representations. This makes it possible to perform interesting analogical reasoning using simple vector arithmetic. For example, the result of the vector calculation: vec("king") - vec("man") + vec("woman") is closest to vec("queen"). Also, vec("russia") + vec("river") is close to vec("volga\_river"). 

Furthermore, \cite{resource-word2vec} extends the previous model to include vector representations for phrases as well. They based their work on the insight that idiomatic phrases like "Boston Globe" and "Air Canada" can be semantically understood well by combining the individual words within the phrases. Therefore they treated phrases as individual tokens (just like words), but limiting their vocabulary to only those phrases which appear frequently together, and infrequently in other contexts. They train vector embeddings of dimensionality 300 for these words and phrases, and release the on the internet for public use \footnote{https://code.google.com/p/word2vec/}.

\section{Deep learning in image and text multimodal models}

There has been a lot of progress in multi-label classification problem of associating images with individual words or tags. However, the more challenging problem of associating images with complete natural sentences has only recently started to gain attention. The research in this area has focused primarily on two tasks, namely:
\begin{enumerate}
\item Mapping images and sentences into a combined space
\item Generating descriptions of images in terms of complete and natural sentences.
\end{enumerate}
The first poses it as an information retrieval problem, while the later treats tit as a natural language generation problem.

\subsection{Introduction}
\cite{model-seminal_paper-hodosh} have written one of the seminal papers in this field, providing an interesting discussion on the problem statement, an in-depth comparison of the available data sets (including what kind of data is required for good modeling), an analysis of modeling techniques employed in the early stages of this field, and a detailed discussion of the various evaluation metrics used in different previous related works.

\cite{model-seminal_paper-hodosh} go into the philosophy of image description by arguing that there are three different kinds of image descriptions:
\begin{enumerate}
\item \textit{Conceptual} descriptions identify what is being depicted in the image. They are concerned with the \textit{concrete} descriptions of the depicted scenes and entities, their attributes and relations, as well as events they participate in.
\item \textit{Non-visual} descriptions provide additional background information that cannot be obtained from the image alone, e.g about the situation, time or location in which the picture was taken.
\item \textit{Perceptual} descriptions capture low-level visual properties of images, e.g if it is a photograph or a sketch, or what colors or shades dominate.
\end{enumerate}
They argue that out of these three, \textit{conceptual} descriptions are the most relevant for image understanding tasks. They observe that using user generated captions uploaded with images in popular image-sharing websites (such as Flickr.com) do not serve as good training data  because people tend to provide information that could not be easily obtained just from looking at the image itself. For example, the kind of description our models require is \textit{"Three people setting up a tent"} while people tend to provide captions like \textit{"Our trip to the Olympic Peninsula"}. Hence they establish a need of data collected purposefully for this specific task.

Most of good quality data sets are collected visa crowdsourcing (for example using Amazon Mechanical Turk) in which multiple descriptive captions are assigned to each image. Pascal1K \cite{resource-PascalVOC}, Flickr8K \cite{resource-Flickr8K}, Flickr30K \cite{resource-Flickr30K}, and MS-COCO \cite{lin2014microsoft} are examples of such good quality data sets.

\bigskip
Some of the earliest work in this field used shallow learning techniques and fixed (as opposed to learn-able) image and text features. \cite{makadia2010baselines,ordonez2011im2text} use nearest neighbour search \ac{kNN} for image annotation and image description respectively. On the other hand, \cite{model-seminal_paper-hodosh} use \textit{kernel canonical correlation analysis} \ac{KCCA}. KCCA \cite{bach2003kernel} is technique which takes training data consisting of pairs of corresponding items drawn from two different modalities and finds maximally correlated linear projections of each set of items (by first mapping the original items into higher-order spaces) into a newly induced common space. Similarly, popular shallow image features include SIFT descriptors \cite{lowe2004distinctive} and simple bag if words (BoW) kernel.

\subsection{Evaluation Metrics}

Since image description is a subjective task, the ideal setting for evaluating a system would be averaged human judgement. However, since human judgement is expensive and slow, there have been a number of metrics employed in evaluating these systems. These metrics can be divided into two categories:
\begin{itemize}
\item metrics for text generation systems.
\item metrics for ranking systems.
\end{itemize}

\textbf{BLEU} \cite{papineni2002bleu} and \textbf{ROUGE} \cite{lin2004rouge} scores are popular metrics in the automatics image description generation systems. Originally, they are standard metrics for machine translation and summarization respectively, but have been used to evaluate multiple caption generation systems \cite{model-NIC-samy,ordonez2011im2text,gupta2012choosing}. Given a caption $c$ and an image $i$ with a set of reference captions $R_i$, the BLEU score of a proposed image-caption pair $(i,s)$ is based on the n-gram precision of $s$ against $R_i$, while ROUGE is based on corresponding n-gram recall. If $c_s(w)$ is the number of times word $w$ occurs in $s$, they are defined as:

$$BLEU(i,s) = \frac{\sum_{w \in s} min(C_s(w), max_{r \in R_i} c_r(w))}{\sum_{w \in s} c_s(w)}$$
$$ROUGE(i,s) = \frac{\sum_{r \in R_i} \sum_{w \in R} min(C_s(w), c_r(w))}{\sum_{r \in R_i} \sum_{w \in r} c_r(w)}$$

\cite{model-seminal_paper-hodosh} try to compare BLEU and ROUGE scores against human judgements, and examine to what extents do these both agree. Based on results they question these metrics' usefulness for evaluating caption generation systems. \cite{reiter2009investigation} also argue that they are more useful as metrics for fluency, but poor measures of content quality of language generation. However, unless a more suited metric is devised, these score are continue to be used for evaluating modern caption generation systems.

\bigskip
Next, we will touch upon metrics which can be used to evaluate the quality of a ranked list in information retrieval tasks. \textbf{Recall@K} (R@K) is the percentage of test queries for which a model returns the correct result among the top $k$ results. It is especially useful in the context of search where a user may be satisfied with the first $k$ results containing a single relevant item. Conversely, \textbf{median rank} is equal to the value of $k$ for which the $R@K$ is equal to 50\%. The $k$ in $R@K$ typically varies between $k=1,10$. \cite{model-seminal_paper-hodosh} consider $k=1$ as a very strict threshold and, after comparing it with human judgement, view it as a lower bound on actual performance.

In our systems queries can have multiple (variable number) of relevant answers since each test image may be associated with multiple relevant captions, and each test caption may deem fit for multiple images besides the one it was originally written for. \textbf{R-precision} \cite{} is the metric of choice in these conditions, since it is a single number which allows us to rank models according to their overall performance (no threshold like $k$) The R-precision of a system with query $q_i$ and known relevant results $r_i$ is defined as its precision at rank $r_i$. In simpler terms it is the percentage of relevant items among the top $r_i$ responses returned by the system.

\subsection{Image and text mapping}

\cite{model-devise-samy} were one of the first ones to use deep learning in ranking images with text.They call their model \textit{DeViSE} or \textit{deep visual-semantic embedding}. Their original task was to improve performance of their image classification system for large number of object categories and  for labels on which the visual system was not trained (\textit{zero shot prediction}). They propose to leverage information from textual information of image to improve their object classification performance.

They begin by pre-training an efficient deep convolutional neural network (CNN) for visual object recognition, based on the architecture by \cite{vision-krizhevsky}. In parallel, they pretrain a simple neural language model well-suited for learning semantically-meaningful vector representations of individual words (\textit{word embeddings}) using skip-gram text modeling architecture \cite{resource-word2vec-related,resource-word2vec}. Following this, they construct the DeViSE model by chopping of the top layers of the CNN and re-training it to predict the word embedding vector of the corresponding image label.

They used hinge margin ranking loss criterion in the second phase of their training, and observe significant improvements over using $L_2$ loss criterion. Although they never trained or tested their system for natural image descriptions, but they did influence a lot of work in this field. \cite{model-defrag-feifei} implemented their model for image and sentence mapping to compare performance against their own system.

\bigskip
\cite{model-week_annotation-hodosh} use CNN for modeling images in their image based sentence retrieval system. They first embed image and sentence into a common space and then use it to rank the pair. They perform a comparison between using 4,096 CNN activations (trained on ImageNet \cite{resources-imagenet-feifei}) as image features versus KCCA with 4,000 dimension fixed features. There was a reported 9.5\% improvement in Recall@10 for CNN over KCCA, even when the CNN activations remained fixed and it was not fine-tuned on image-text data set.

\bigskip
Unlike previous work, \cite{model-defrag-feifei} go on a finer level and map fragments of images (objects) and fragments of sentences (dependency tree relations) into a common embedding space. Their model works for bi-directional retrieval: image given a text query, and text given an image query. They interpret an image being made up of multiple entities, and therefore propose to break it down into more manageable fragments.\ac{ANN}s are used to compute vector representation of these image and sentence fragments in a multimodal embedding space, and the dot product between a pair of these vectors (one image fragment and one sentence fragment) are interpretted as a compatibility score. The global image-sentence compatibility score is computed as a fixed function of their fragments.

Their image model comprises of a Region convolutional neural network (RCNN) which detects objects in the images and returns their bounding boxes. For image fragments they use the top 19 locations detected by the RCNN and the complete image, so each image is broken down into 20 fragments. Following this another CNN is applied on each of these fragments to return a 4,096 dimensional embedding vector representing the image fragment. The architecture of this CNN closely resembles that by \cite{vision-krizhevsky}. On the other hand, the sentence fragments are considered as edges of the sentence's dependency tree. Therefore, each sentence fragment consists of two words which are joined in any stage of the dependency tree. Each word in the dictionary of 400,000 is represented using 1-of-k encoding, and vector embeddings for the words are obtained through an unsupervised objective and fixed throughout the training. The sentence fragment score is calculated using these word embeddings as well as separate embeddings which represent the type of relation between the words.

They plot a matrix where the rows represent the image fragments and the columns represent the sentence fragments. Each element of the matrix (or box) shows the dot product score between those two multimodal fragments. They define two kinds of objective functions to train their models:
\begin{itemize}
\item \textbf{Fragment alignment objective:} This objective explicitly learns the representation of sentence fragments in the visual domain. It encodes the intuition that if a sentence fragment is contained in an image, at least one of the boxes should give a high score with that sentence fragment, while all other boxes corresponding to images which do not contain this sentence fragment (in their descriptions) should have a low score with this fragment. It also favours that there should be at least one high scoring box in each column of the matrix (meaning every sentence fragment should be matched by at least one image fragment). 
\item \textbf{Global ranking objective:} This objective tries to enforce that the image-sentence similarities are consistent with the ground-truth. First the global similarity score is computed by averaging the pairwise fragment scores. After this the entire system is trained in an end-to-end fashion with the margin ranking loss criterion.
\end{itemize}
The model improves performance when using a combination of these two objective functions. They also report that dividing image in fragments has performance improvement versus treating the entire image as a single fragment, and that dependency tree sentence fragments perform consistently better than bag of words (BoW) and bigram features. Finally they found that fine-tuning the image score calculating CNN on image-text data further improves results.

\bigskip
\cite{model-SDT_RNN-socher} go one step ahead and use the entire sentence dependency tree - as opposed to only its edges as sentence fragments - to model the sentences using DT-RNN (dependency tree recursive neural networks). They argue this is important for accurately representing complicated sentences in the visual domain. They test their model against other recurisive and recurrent neural networks, KCCA and BoW baseline, concluding that DT-RNN out performs all of these in the task of image and sentence mapping. DT-RNN also give similar vector representations to multiple captions which describe the same image, adding more weight to their model. The DT-RNN is different from other previous recursive neural network models \cite{text-RNN-socher} which are based on constituency tree (CT-RNN). They report that the sentence vectors computed by DT-RNN are more apt at capturing the meaning of the sentence in terms of "visual representation". Moreover DT-RNN vector embeddings are more robust to changes in syntactic structure or word order as opposed to CT-RNNs or recurrent neural networks. The final sentence embedding is a vector of length 50.

For the image side, they train a deep CNN first using unsupervised objective: reconstruct the input keeping the neurons sparse, followed by supervised learning on classifying 14 Million images of ImageNet into 22,000 categories. The CNN used was particularly large, with 1.36 billion parameters, and they achieved 18.3\% precision@1 on the full ImageNet data set. Following the CNN training, they chop off the last layer to get 4,096 dimensional vector embeddings for the images. 

During the multimodal mapping, the 4096 they transform the image representation vector to the same size as sentence vector. Following this - like other similar works -  they take a dot product of these to produce a compatibility score and back propagate error using margin ranking loss criterion. However, the 4,096 dimensional image vector and 50 dimensional sentence vector are fixed and are not updates during this phase. They report improved results over the Pascal1K data set \cite{resource-PascalVOC} compared to bag of words (BoW), CT-RNN, recurrent ANN and KCCA models.

\bigskip
\cite{model-brnn-feifei}  build upon their prior work on image text matching using fragments of image and sentence \cite{model-defrag-feifei} (discussed above), add a new features to increase the modeling capacity of their model. Their initial approach translated words directly into vector embeddings, and did not consider word ordering and context. To address this problem they use a bi-directional recurrent neural network (BRNN) to model the input text sequence and convert it into a complete sentence embedding. They report significant improvements using this approach.

\bigskip
Taking a somewhat different approach, \cite{model-multimodalDBM-ruslan} use multimodal deep boltzmann machines (DBM) for this task. The model learns joint probability density over the space of multimodal inputs. Missing modalities can be filled in by sampling from the conditional distributions over them. For example we can learn the joint probability $P(v_{img},v_{txt}\mid\theta)$ ,where $v_{img}$ is the vector representation of an image and $v_{txt}$ is the vector representation of a text sentence. Once this distribution is estimated we can draw samples from the conditional probabilities to fill in the missing modalities: drawing samples from $P(v_{txt}\mid v_{img},\theta)$ to give the missing text sentence, and vice versa.

A multimodal DBM is an extension of an DBM to model multimodal inputs. A DBM is formed by stacking together RBMs, in other words a multilayer RBM, which are usually trained in a greedy layer-wise strategy. They construct two independent DBMs: an image-specific DBM and a text-specific DBM. Next these two are connected together via another RBM layer to construct the multimodal DBM. In other words, the outputs of the two DBMs are connected to a another layer of binary hidden units on top of them, called the \textit{"joint representation"}. The intuition behind this model is that each data modality has very different statistical properties which make it difficult for a single hidden layer model to directly find correlations across modalities. This difficulty is overcome by putting layers of hidden units between the inputs of both modalities (the separate DBMs)

They evaluate their model on information retrieval for multimodal and unimodal queries. In multimodal query, the aim is to give higher similarity score to the image and text pairs belonging to the same instance, over false pairings. While in unimodal query, either text or image is provided, and the model predicts the missing modality out of a pool of possible options.

\subsection{sentence generation from images}\label{sentence_generator}

Now we will briefly touch upon the related task of generating natural sentences from images. Although it is not the focus of our project, but it is still to interesting to see the approach taken in that area, especially since there has been a growing interest in it recently. \cite{model-multimodalNLM-ruslan} present a model in which they use a convolutional neural network (CNN) for this task. The CNN has four input pathways, one for an image and the remaining three for words. The idea is that, given an image and three previous words, the model learns to predict the next word in the sequence. In this way they learn a joint "multimodal language model".

\bigskip
Recurrent neural networks (RNN) are quite popular for text generation, and so many researchers use them  in this task, albeit in different settings \cite{model-brnn-feifei,model-NIC-samy}. \cite{model-NIC-samy} are influenced by modern ANN based machine translation systems, and they employ a encoder-decoder type architecture for their model. Specifically, they use a CNN as an encoder and an RNN as a decoder. The CNN encodes the input image in vector space feature embeddings which are are input to the RNN. The RNN takes the image encoding as input and the word generated in the current time step to generate a complete sentence one word at a time. Special words have been marked to tell the model that it is the starting word and for predicting the finishing word. They also use long short term memory (LSTM) inside the RNN so that it has some memory of older generated words. They train their model to directly optimize the log likelihood of the target sentence given the input image.

Using a similar approach, \cite{model-brnn-feifei} also use an RNN trained on multimodal inputs. For each input image feature vector, and the current generated word (starting at a fixed word) it learns to predict the next word in a sequence (ending at a fixed word).
\cleardoublepage
\chapter{Resources}\label{ch:resources}
We employed various data sets and resources in our research and experimentation. Below we will provide a brief description for  them. Most of the high quality data set is collected through crowd sourcing methods, labeled by human input. Although some large-scale automatically generated data sets exist (for example using user provided annotations when uploading images on photo-sharing websites). However, \cite{model-seminal_paper-hodosh} makes a convincing case about the lack of pertinence of these captions with the objectives of our task, and hence we do not use them (like many researchers in the domain).

\section{Flickr8K}
\textbf{Flickr8K} \footnote{http://nlp.cs.illinois.edu/HockenmaierGroup/8k-pictures.html} is a data set comprising of 8,092 images collected from the Flickr website \cite{resource-Flickr8K}. Each image has five captions along with it, which describe the contents of the image. The images in this data set focus on people or animals (mainly dogs) performing some action. The images tend not to contain well known locations, but are manually selected to depict a variety of scenes and situations. The images are captioned by human captioners (five for each image) using crowdsourcing via Amazon's Mechanical Turk \footnote{https://www.mturk.com/}.

In order to avoid grammar mistakes, the captioners (who were only from US) had to pass a brief quality control test based on spelling and grammar. Following this, they were asked to write sentences that describe the depicted scenes, situations, events and entities (people, animals, other objects). Multiple captions were collected to model the inherent variability of different humans in describing the same image. As a consequence, the captions of the same images are often not direct paraphrases of each other: the same entity or event or situation can be described in multiple ways (man vs. bike rider, doing tricks vs. jumping), and while everybody mentions the bike rider, not everybody mentions the crowd or the ramp.

It is worth mentioning that when accessing the data set, we found out that several of the images had been removed from Flicker (the authors gave links redirecting to the original images \footnote{http://nlp.cs.illinois.edu/HockenmaierGroup/8k-pictures.html}) so we could only get download 6,793 out of the total 7,678 images. Therefore, although we can use to judge the performance of our algorithm on the data set, we cannot use it in comparison to previous work.

\section{Flickr30K}
\textbf{Flickr30K} \footnote{http://shannon.cs.illinois.edu/DenotationGraph/} is an extension over the Flickr8K data set, comprising of 31,783 images collected by \cite{resource-Flickr30K}. Similar to its counterpart, each image is associated with five captions written by humans using the Amazon's Mechanical Turk. The images consists of everyday activities, events, and scenes.

It is important that the annotators are not familiar with the specific entities and circumstances depicted in them, to avoid that they give overly specific annotations. For example the annotations "Three people setting up a tent" are more relevant to the task as compared to "Our trip to the Olympic peninsula" (for the same image - the latter is likely an annotation given by a person familiar with the significance of the image). Moreover different annotators use different levels of specificity, from describing the overall situation (performing a musical piece) to specific actions (bowing on a violin).This variety of descriptions associated with the image is necessary to induce similarities between expressions that are not trivially related by syntactic rewrite rules.

\section{OverFeat}\label{resource:overfeat}
\textbf{OverFeat} \footnote{http://cilvr.nyu.edu/doku.php?id=software:overfeat:start} is a publicly available visual feature extractor based on the convolutional neural network submitted by \cite{resource-overfeat} to ILSVRC-2013 large-scale object recognition competition. At the time of its release it ranked $4^{th}$ in classification, $1^{st}$ in localization, and $1^{st}$ in detection tasks of ILSVRC-13 data sets. We use the $accurate$ version of their model which has 144 million free parameters and 5.4 billion connections, and reaches performance of 14.18\% on the competition validation set. Their network architecture is similar to the architecture by \cite{vision-krizhevsky}, with the addition that it is trained on images at multiple scales, and it has improved inference step.

\section{Word2vec}
\textbf{Word2vec} \footnote{https://code.google.com/p/word2vec/} is a  publicly available tool which provides an efficient implementation of learning continuous bag of word, and skip-gram based vector representations for words. The model and implementation is based on the work of and released by \cite{resource-word2vec}. In addition to the implementation, they also provide vector representations of words and phrases which they learned by training this model on Google News Dataset (about 100 billion words). These are 300-dimensional vectors for 3 million words and phrases. An interesting feature of these vector representations are that they capture linear regularities in the language. For example the result of the vector calculation: vec("Madrid") - vec("Spain") + vec("France") is closest to vec("Paris").

\cleardoublepage
\chapter{Research Methodology}\label{ch:methodology}

This section will provide an overview of research methodology and time line of experiments we conducted. Most of the experiments mentioned in this chapter are not directly involved in our final results, and so the reader can skip over this chapter if they wish to. Any material which is deemed necessary to understand our model will be reiterated in the next chapter. The reason for this chapter's inclusion is to make the reader understand how our experience with the problem influenced the way we approached the task (including the set backs we faced), and how we arrived to our final model.

\bigskip
We decided to treat the problem of image descriptions as an information retrieval task, and decided to follow a ranking approach for matching images and sentences. This decision was influenced by prior work at NEC relating to SSI (supervised semantic indexing) \cite{text-SSI} which is a system for document retrieval based on a query sentence. The basic idea is to generate vector embeddings for different texts (query and document) which contain semantic information about the original text. These embeddings act as abstraction of the content in text, and can be used for similarity comparisons between different texts.

SSI is quite similar to our problem, with the difference that our inputs are multimodal. In SSI both query and result are of textual nature, while for us, one is text and the other is image. The presence of multimodal inputs makes our task much more challenging, therefore we would need to change the architecture to cater for this complexity.

\bigskip
We are interested in making our model bi-directional, meaning that it would be able to retrieve an image for a textual sentence query (\textbf{image search}) and a sentence for an image query (\textbf{image annotation}). Following this goal, we aim to make our model different in two significant ways from the SSI architecture:
\begin{enumerate}
\item We will divide our network into a \textbf{visual model} and a \textbf{textual model}. The visual model will be fed with images and the input for textual model will be images. The goal of each model will be to efficiently extract semantic features from their respective input signals.
\item We will make the overall architecture deeper. Since both our inputs belong to different modalities and differ in statistical properties, transforming them into a common embedding space would be more challenging and we would need the modeling power of deeper networks (with more non linear layers)
\end{enumerate}

We follow the above guidelines to design our model. For the visual model we use a CNN (convolutional neural  network) initialized from random weights. For the textual model we use a simple MLP with a single hidden layer, and we treat the sentence as a bag of words (BoW). We trained this model in an end-to-end fashion with margin ranking criterion. The objective of the criterion is to maximize the distance between a positive and a negative example (giving higher score to the positive example and lower to a negative). A negative example is generated by making a random false pairing with a sentence and image from our training examples (whose ground truth we know). At each iteration we give a positive example and a randomly generated negative example. The gradient is back propagated through the entire network to train it in an end-to-end manner.

\begin{figure}[ht]
\centering
\includegraphics[width=0.7\textwidth]{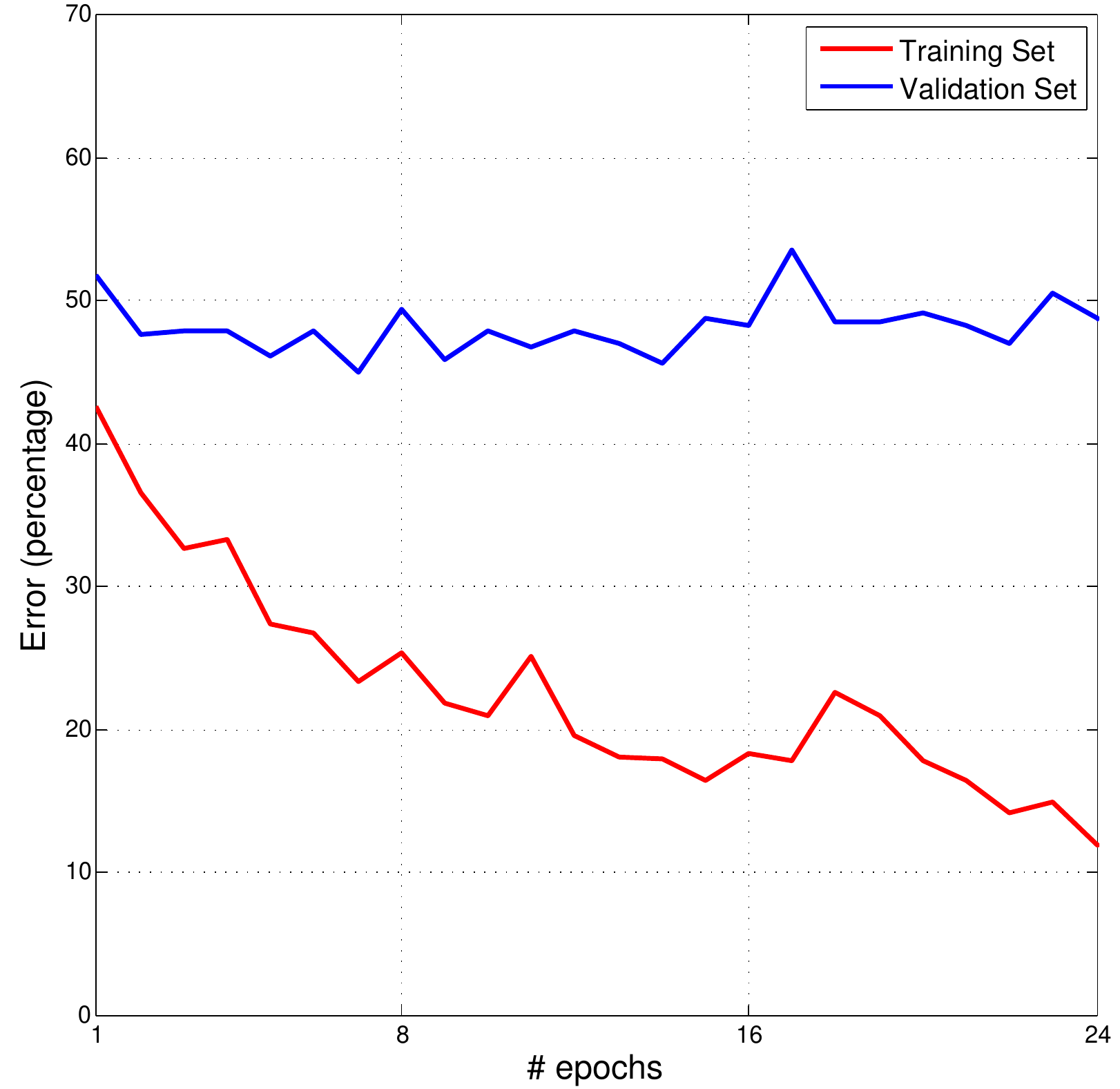}
\caption{Our model fails to perform on \textbf{Flickr8K} data set. Error is the percentage of samples for which the model gives higher score to the positive sample over the negative sample. 50\% error specifies random performance.}
\label{fig:flickr8k_error_bad}
\end{figure}

\bigskip
However the model failed to perform well when trained on Fickr8K data set. The training error decreased with increasing number of epochs, but the error on validation set error seem to remain very close to random, as illustrated in Figure \ref{fig:flickr8k_error_bad}

We tried to analyze our results and investigate into a possible reason for failure. Changing different hyper parameters like learning rate, hidden layers and hidden units did not seem to produce any significant effect. Since SSI has a similar architecture and it seems to perform well on only textual input data, so we could think of three possible reasons of the models failure:
\begin{enumerate}
\item More training data required as the model seems to start over-fitting.
\item our CNN not powerful enough to produce good visual representations.
\item Our model architecture not well suited for the task.
\end{enumerate}

\begin{figure}[ht]
\centering
\includegraphics[width=0.7\textwidth]{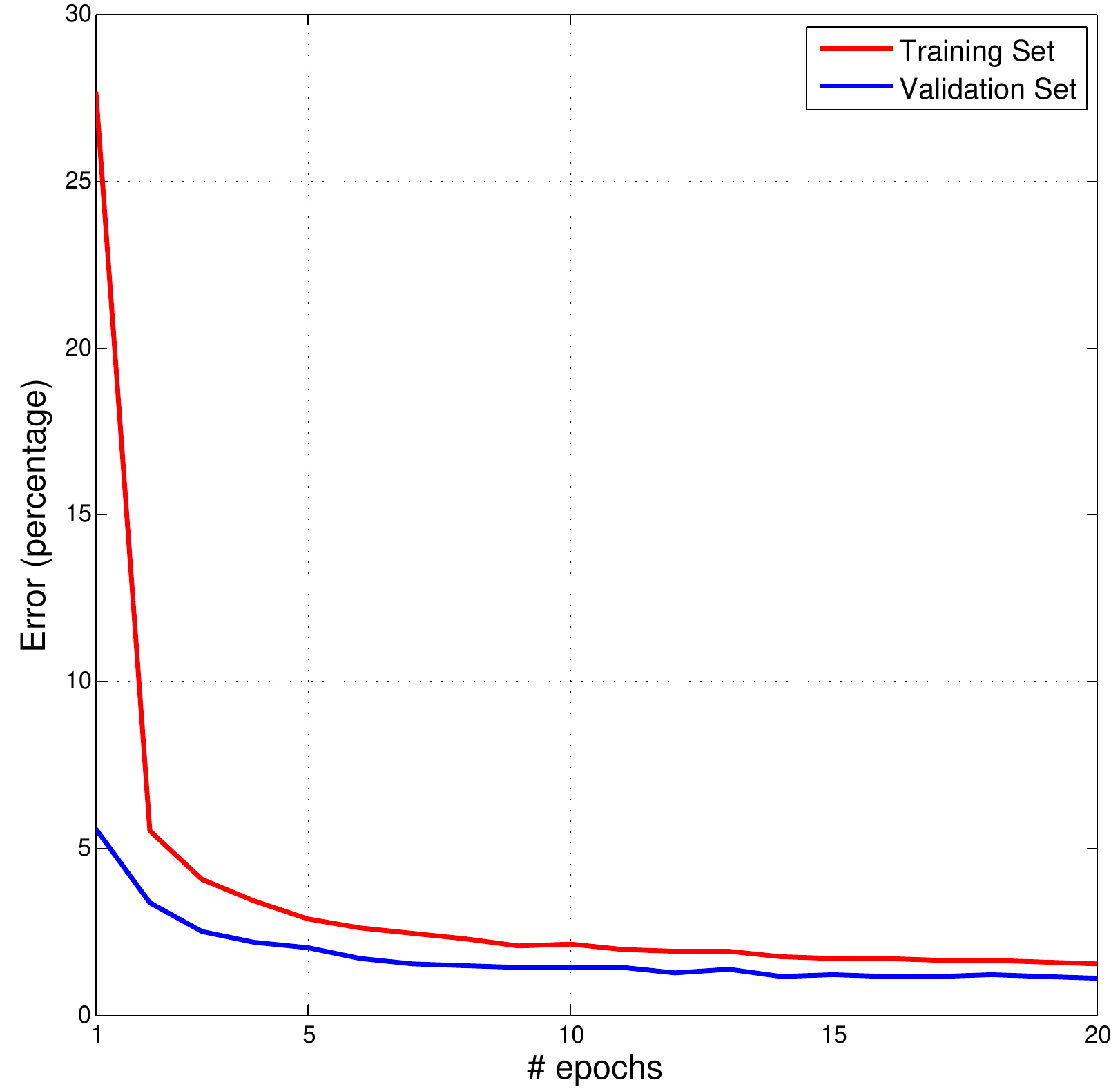}
\caption{Our randomly initialized CNN performed well on classifying \textbf{MNIST} data set. Error is the percentage of samples classified correctly (over 10 categories).}
\label{fig:mnist_classification_good}
\end{figure}

\bigskip
The CNN we used was made out of 5 convolutional blocks followed by 2 fully connected layers. Each convolution block had a convolution layer, non-linearity layer, sub-sampling layer, dropout layer and (subtractive) normalization layer. We decided to first verify if the CNN was working as expected by testing it in isolation. We did this by using it for a much simpler task: classification of handwritten digits from the MNIST data base. The model seemed to perform well without tweeking much parameters, verifying that it was at least functioning properly (refer to Figure \ref{fig:mnist_classification_good} for classification results).   

\begin{figure}[ht]
\centering
\includegraphics[width=0.7\textwidth]{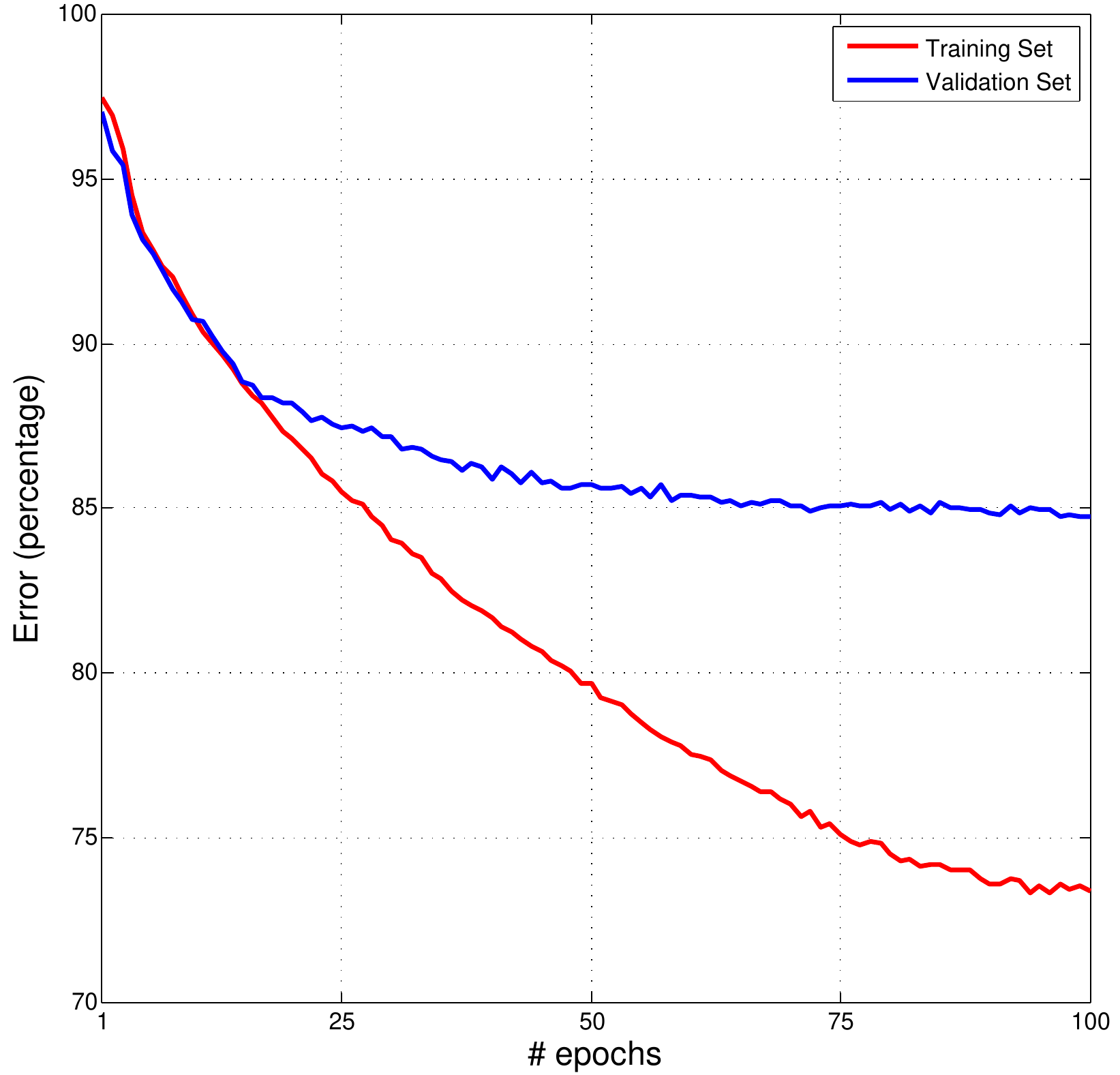}
\caption{Our randomly initialized CNN failed to perform well on classifying \textbf{Caltech256} data set. Error is the percentage of samples classified correctly (over 256 categories).}
\label{fig:caltech_classification_bad}
\end{figure}

Following this we tried to use it on a more complicated task: classification of objects into 256 categories from Caltech256 data set. However, the model failed to perform well on this task, as illustrated in Figure \ref{fig:caltech_classification_bad}. We concluded that the reason for bad performance on the Flickr30K ranking task and Caltech256 classification task was due to the same reason. Our CNN was not able to model the complicated data set well enough. In other words, since the number of images was relatively limited and the amount of variation in the images was rather high (especially true for Flickr30K data set), the CNN was not able to generalize, and model the relevant statistical properties well enough. Therefore we concluded that we needed to pre-train our network with larger data set before attempting to train it on these problems. The advantage of pre-training would be that the model would have very good initialization, and would be able to extract good visual features from the images which could be used for our task.

\begin{figure}[ht]
\centering
\includegraphics[width=0.55\textwidth]{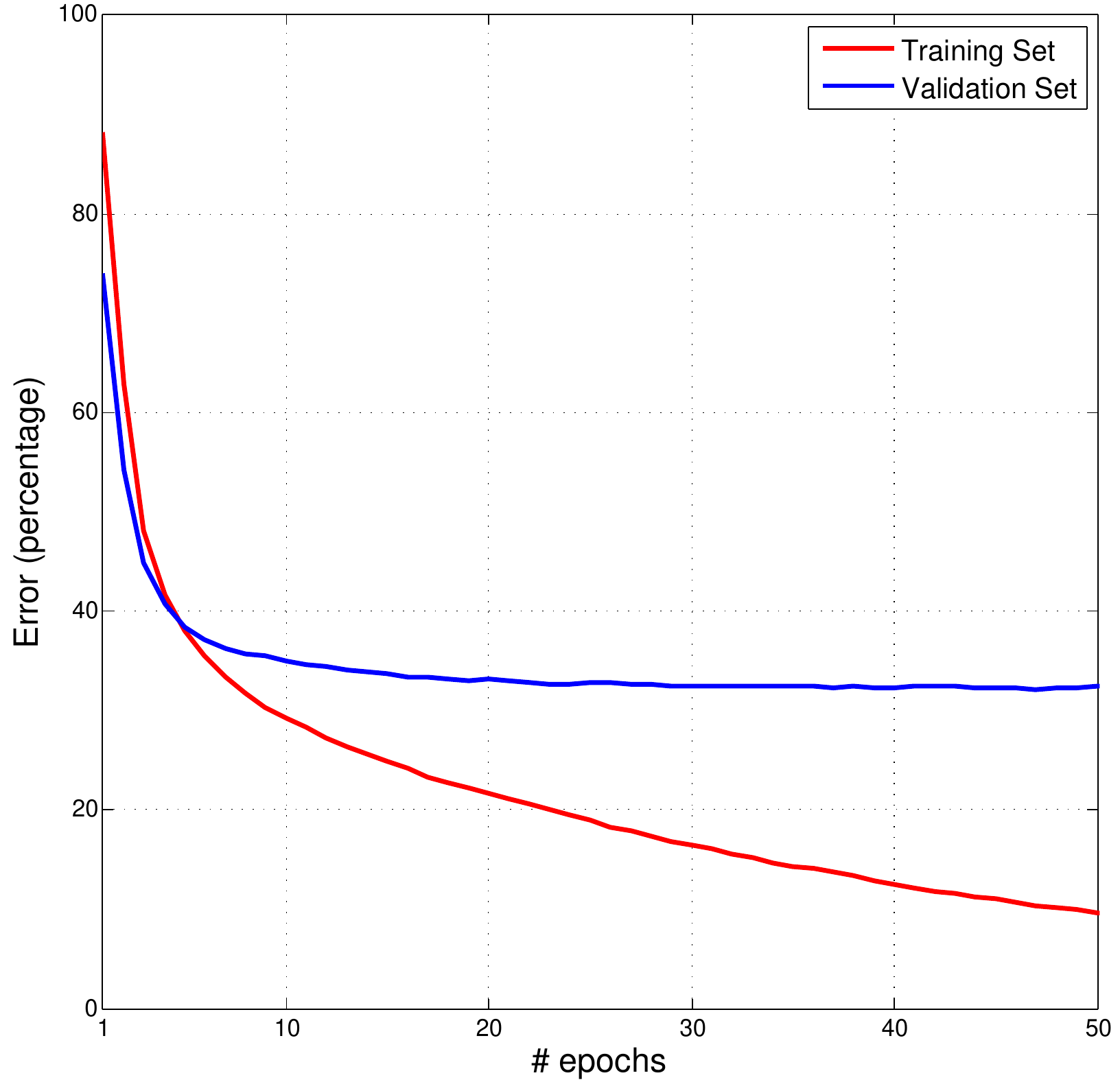}
\caption{OverFeat CNN performed well on the classification task on \textbf{Caltech256} data set. Error is the percentage of samples classified correctly (over 256 categories).}
\label{fig:caltech_classification_good}
\end{figure}

Instead of wasting scarce time to pretrain our network from scratch on a much larger data set (Which could take weeks), we decided to use a publicly available CNN: \textbf{OverFeat} \footnote{http://cilvr.nyu.edu/doku.php?id=software:overfeat:start}. For details of this model refer to Chapter \ref{resource:overfeat}. However, before plugging OverFeat network directly into our original task we decided to first test it on classifying Caltech256 data (on which our CNN failed). We chopped off the last softmax layer of OverFeat and replaced it with a small neural network (1,000 unit linear layer, $relu$ non-linearity, another 1,000 unit linear layer, and finally a 256 unit softmax layer). The parameters of CNN were fixed and not fine-tuned on the new data, and the parameters of the small network on top were only updated. This model managed to classify Caltech256 significantly better than our original randomly initialized CNN (classification results in \ref{fig:caltech_classification_good}). 

\begin{figure}[ht]
\centering
\includegraphics[width=0.55\textwidth]{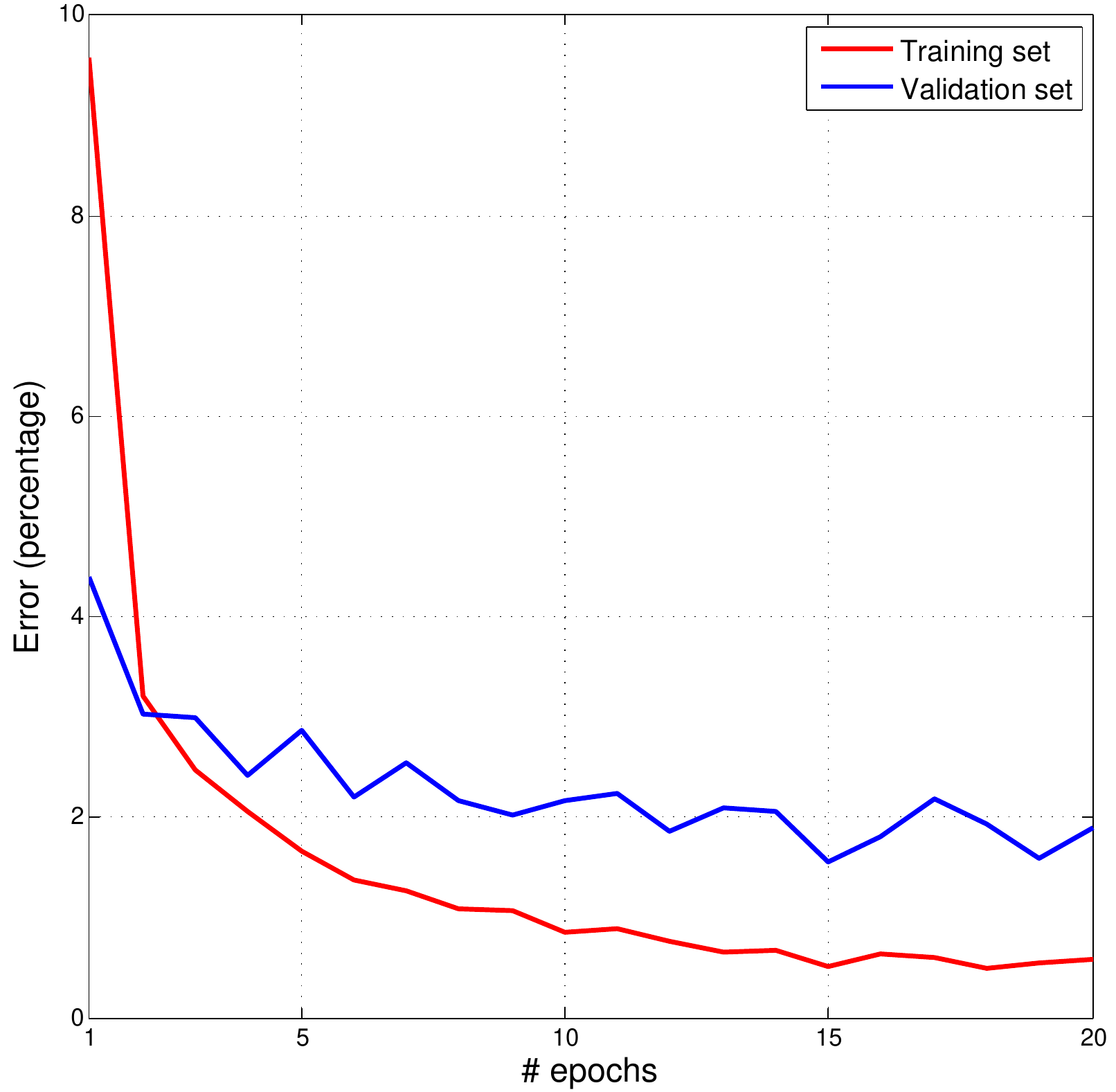}
\caption{OverFeat CNN performed well on the ranking task on \textbf{Caltech256} data set. Error is the percentage of samples for which the model gives higher score to the positive sample over the negative sample. 50\% error specifies random performance.}
\label{fig:caltech_ranking_good}
\end{figure}

We also decided to do a small toy experiment to test the image and text multimodal ranking approach based on Caltech256 data set. For the textual input we used the class labels corresponding to each image, which are incorporated easily since it is a simple bag of words model. Often the label were hyphenated like "baseball-bat" and "bear-mug"; in these cases we decided to break  into individual words (so images belonging to these images would have two word inputs instead of the usual one). The total vocabulary size was 322 words. To our satisfaction, the this toy ranking task also performed well, reinforcing our confidence in the model. The results are shown in \ref{fig:caltech_ranking_good}. Therefore, we finally decided to do further experiments with OverFeat CNN model.

\cleardoublepage
\chapter{Proposed Model}\label{ch:proposed_model}

\section{Abstract Architecture}
Our model is inspired by the previous work at NEC of SSI (supervised semantic indexing) \cite{text-SSI} which is a textual retrieval system. We aim to extend this model to multimodal inputs: in particular images and text. We will focus on bidirectional retrieval. In other words, retrieving images based on a sentence query (\textbf{image search}) and retrieving sentences based on an image query (\textbf{image annotation}). The sentences are complete and natural, constructed by human annotators, as opposed to individual and isolated tags.

\begin{figure}[ht]
\centering
\includegraphics[width=0.7\textwidth]{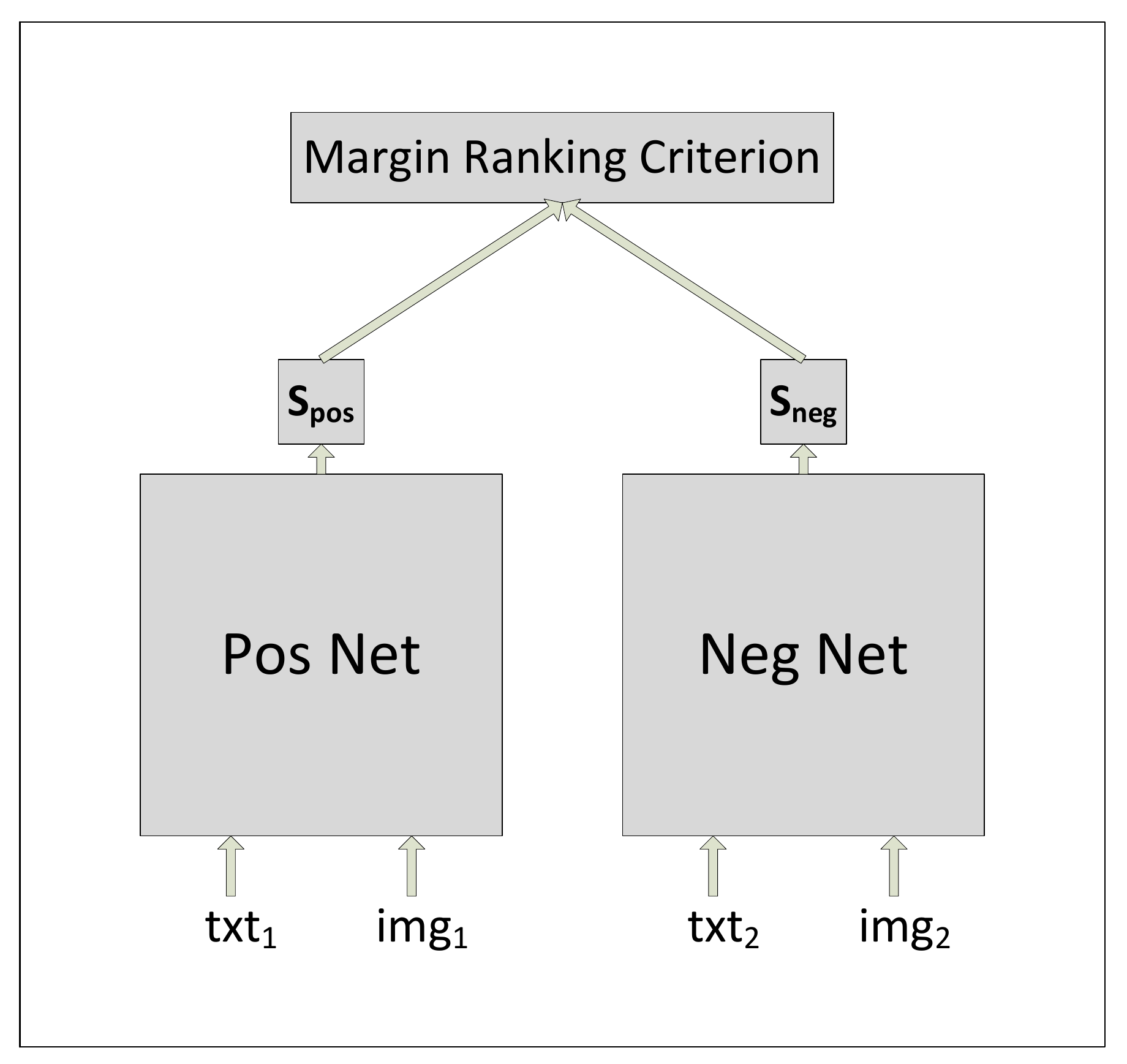}
\caption{Overall abstract level architecture of our complete multimodal retrieval model. $Pos Net$ and $Neg Net$ share weights. They generate a score $s$ based on similarity (and semantic relevance) between a given pair of text ($txt$) and image ($img$) input}
\label{fig:flowchart_overall_arch}
\end{figure}

\bigskip
The overall abstract-level architecture of our complete model is illustrated in Figure \ref{fig:flowchart_overall_arch}. The two sub-models $Pos Net$ and $Neg Net$ have identical architecture and they share weights. For a pair of text and image input, they calculate the score $s$, which is a measure of the similarity between the corresponding text and image (based on cosine distance). Here $S_{pos}$ is the similarity score of the positive input pair and $s_neg$ is the similarity score of negative input pair. Positive pair means that the sentence positively describes the image (As determined by human judgement). Negative pair means that the image and sentence are not related, and the pair is generated by simply replacing the sentence (or the image) in a positive pair by another random sentence (or image).

\bigskip
At each iteration we simultaneously present a positive pair and a negative pair to the $Pos Net$ and $Neg Net$ respectively. The objective criterion penalizes if  $S_{pos}$ is not greater than a certain threshold above $S_{neg}$. Intuitively it means that the system is trained to give high score to  similar pair or sentence and image, and a low score to a dissimilar pair. Therefore the network learns to score semantic relevance between the pair of multimodal inputs.

As discussed previously, in our data set there are five sentences corresponding to each image. Each one of them is labeled by a different person, in order to model natural variance within descriptions. We use all five sentences during the training process, so each image is samples five times during one full epoch (in a random order). We believe it is extremely important to use all five sentences to make the model more robust to variance in natural language.

\begin{figure}[ht]
\centering
\includegraphics[width=0.7\textwidth]{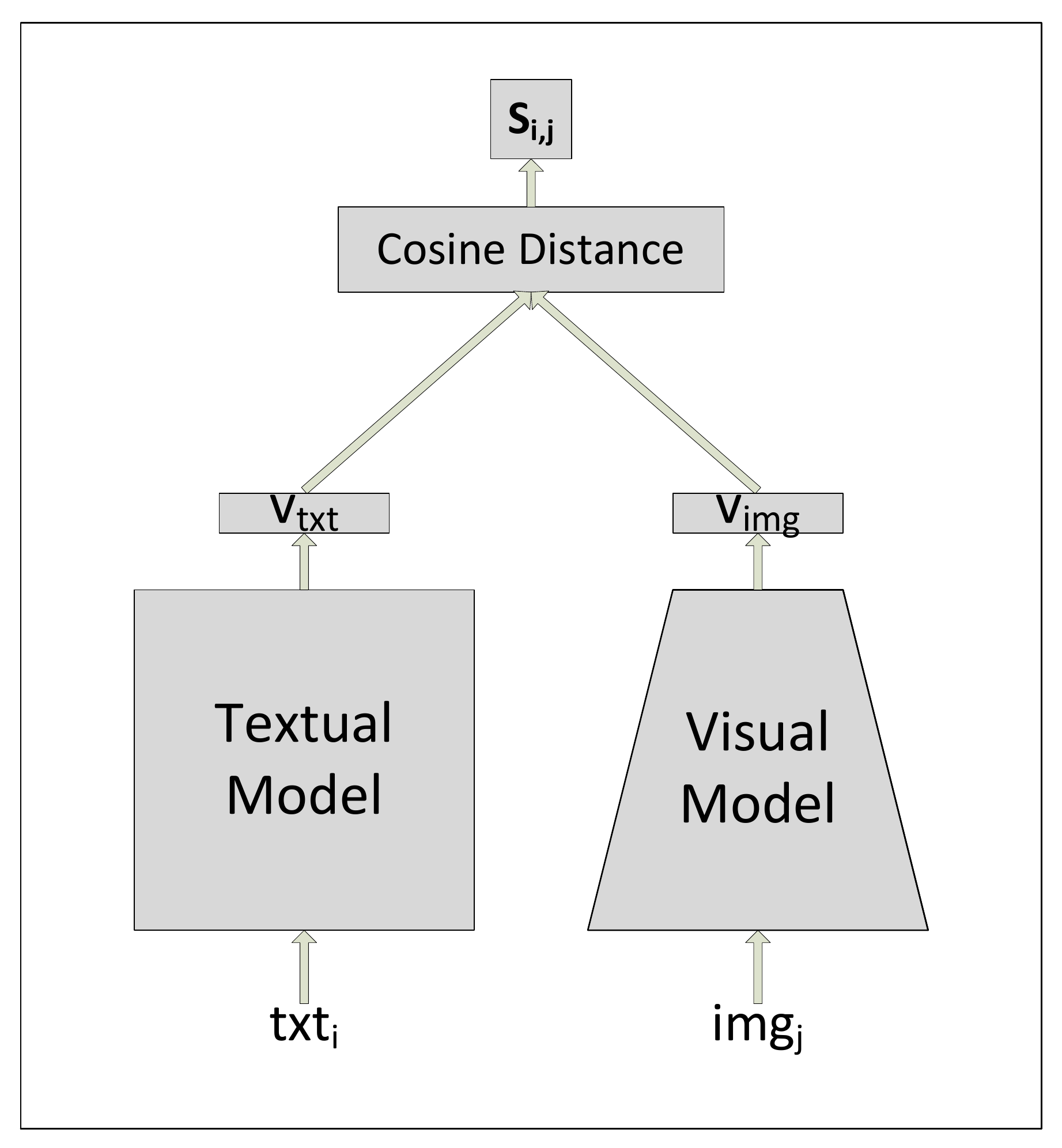}
\caption{Architecture inside the visual-textual model in the form of a Siamese network. Both $Pos Net$ and $Neg Net$ share this architecture. $V_{txt}$ and $V_{img}$ are the fixed dimensional vector embeddings of the textual and visual input respectively. The score $S$ is simply the cosine distance between these two vectors.}
\label{fig:flowchart_siamese_network}
\end{figure}

\bigskip
Next we will go one lower level of abstraction and discuss the inside of the model which makes $Pos Net$ and $Neg Net$. This is illustrated (again in an abstract manner) in Figure \ref{fig:flowchart_siamese_network}. In our system the \textit{textual model} and \textit{visual model} are replaces by ANN (artificial neural networks). Given a sentence and an image, these ANNs are trained to transform them into low dimensional high-quality vector embeddings. $V_txt$ signifies the vector embedding generated from the textual input, and $V_img$ is the vector embedding from visual (image) input.

By high-quality we mean that the vectors are embedded in a common embedding space which encodes high level semantic information about the contents of the image and text while ignoring feature level details (for example that one is text and one is input). The advantage is that once multimodal input is embedded in a common space we can apply standard vector operations to measure their similarity and use it for retrieval task. 

These high quality embeddings are learned through a global ranking objective function. Below we will briefly discuss some of the mathematical details of our model, including the objective function.

$$error_i = max(0, S_{i,neg} - S_{i,pos} + \Delta)$$

$$S_i = \frac{\bar{V}_{i,img} \cdot \bar{V}_{i,txt} }{\left \| \bar{V}_{i,img} \right \| \left \| \bar{V}_{i,txt} \right \|}$$

$$\bar{V}_{i,img} = f_{img}(img_i) $$
$$\bar{V}_{i,txt} = f_{txt}(txt_i) $$

\bigskip
Where, $f_{img}(.)$ signifies the \textit{visual model} and $f_{txt}(.)$ signifies the \textit{textual model}(refer to Figure \ref{fig:flowchart_siamese_network}). Additionally $img_i$ and $txt_i$ are the $i^{th}$ image and sentence (text) inputs respectively. 

The $error_i$ denotes the global objective function. For a given margin threshold ($\Delta$), if the positive input score $S_{i,pos}$ is not at least $\Delta$ bigger than the negative input score $S_{i,pos}$, then the error gradient propagates throughout the architecture. This is shown in the equation below:

$$\frac{\partial error_i}{\partial S_{i,neg}} =\begin{Bmatrix}
0, \mbox{ if } S_{i,pos} \geq  S_{i,neg} + \Delta\\ 
1, \mbox{ otherwise }
\end{Bmatrix}$$

$$\frac{\partial error_i}{\partial S_{i,pos}} =\begin{Bmatrix}
0, \mbox{ if } S_{i,pos} \geq  S_{i,neg} + \Delta\\ 
-1, \mbox{ otherwise }
\end{Bmatrix}$$

\section{Details Visual and Textual models}
Following the deep learning approach we use neural networks for both visual and textual models. This is because ANNs can be trained in an end-to-end manner, optimizing the global objective function, and require minimal preprocessing. Also, as discussed in Chapter \ref{ch:lit_review}, various ANN architectures achieve state of the art results in various machine learning tasks. We hope to reciprocate these results for our task.

\bigskip
For the visual model we use a convolutional neural network (CNN), since they have become ubiquitous with image recognition tasks. As discussed in Chapter \ref{ch:methodology}, we experimented with training a CNN from scratch (with random initialization) on our data set, but that failed to achieve good results. After some experiments we were convinced of the need of pretraining the CNN with large image database like the ImageNet. However, doing that from scratch and achieving state of the art results would take us a very long time, so we decided to explore publicly available CNN models. 

In all our experiments we use the \textbf{OverFeat} convolutional neural network (refer to \ref{ch:resources}), which is a competition winner from ILSVRC-2013. We chop off the last softmax layer of OverFeat which leaves us with 4096-dimensional feature vector. Since this is very deep inside the network, we argue that the features would be high-level and abstract (not spatially dependent and with semantic information) and we can use them directly in out model. Therefore we don't update the weights and parameters of the OverFeat network, and treat is as a feature extractor. Instead we build our own smaller fully connected ANN which takes OverFeat feature vector as input and generates the multimodal image embeddings. Results show good performance with this approach and we stick with it throughout our experimentation.

\bigskip
For the \textit{textual model} we experimented with different architectures and input features. A brief overview of these is given below. Details of these will come in the next chapter when we discuss experiments.
\begin{enumerate}
\item The first and simplest model we try is a fully-connected \textbf{MLP} with unigram bag of words (\textbf{BoW}) features of dictionary size 5,000. The size of the input layer is equal to the vocabulary size, with each index mapping to an individual word. Next there are one or two optional hidden layers with non-linearity, for greater modeling strength. Finally the last layer gives the feature vector $V_txt$ for the entire textual input, and it is of size $n_{emb}$. It is also noteworthy that the first layer would encode information about individual words, and thus we experiment initializing it with good publicly available word embeddings called \textit{\textbf{word2vec}} (refer to Chapter \ref{ch:resources})
\item A major problem with the BoW model is that it looses all sense of structure in the sentence (all words are treated equally and there is no information if a word came before or after another). There are many ways to preserve this syntactic information in the sentence, and in this model we try the simplest approach: \textbf{$n$-grams}. We extend our \textit{textual model} to include \textbf{bi-gram}, \textbf{tri-gram}, and tri-grams with \textbf{skip-grams} as inputs. The vocabulary size (and thus the input layer of MLP) is now increased to 50,000 units (based on the most frequently occurring n-grams). Although our model still as a whole treats each n-gram as a unit inside a bag (with no connection to other n-grams in the sentence). But, there is still some syntactic information present inside the individual n-gram unit, and so we hope to gain some performance due to increased syntactic information in our textual input. We also tried experimenting with using \textbf{tf-idf} features instead of binary word presence features.
\item We also tried experimenting with the supervised semantic embedding (\textbf{SSE}) architecture \cite{text-SSE} as the textual model. SSE uses a specialized ANN architecture to  deal with variable length sentences and temporal dependencies within the text. It has been successfully applied to applications like sentiment analysis from customer reviews at NEC, which led to our motivation in using it.

The architecture for SSE consists of three parts. An embedding layer (\textbf{lookup table}) embeds each word into an $M$-dimensional vector. $K$ kernels are then convolved over the sequence of word embeddings to give $K$ sequences of fixed length (phrase level embeddings). Finally  an averaging layer averages over the length, reducing the sequence to a $K$-dimensional vector, which encodes semantic information about the complete sentence. Further layers can be added on top depending on the required functionality.
\end{enumerate}

\section{Unique training methodology}\label{sec:training_methodology}
As we explained above, we generate a negative sample by fist taking a positive sample and then replacing the sentence (or the image) with another random sentence (or image). So in effect we can take two approaches: replacing the sentence with another sentence (while keeping the image) or replacing the image with another image (while keeping the sentence). It turns out that this choice makes a difference in the evaluation results, so from now on we will refer to this choice as the \textbf{training methodology}. This choice is explained in the mathematical forumalism below (refer to Figure \ref{fig:flowchart_overall_arch} in connection to the equations below)

$$\mbox{I2T approach: } img_1 = img_2 \mbox{ , } txt_1 \neq txt_2$$

$$\mbox{T2I approach: } txt_1 = txt_2 \mbox{ , } img_1 \neq img_2$$

\bigskip
We will discuss the effects of these with empirical results in the next chapter. However, we would briefly mention here that the \textbf{I2T approach} gives higher performance for \textit{image annotation} systems, while \textbf{T2I approach} gives higher performance for \textit{image search} systems. Therefore, this methodology of generating negative samples can be used to specialize the system towards a particular task (although the system is inherently bi-directional and can still perform the other task with reasonable accuracy).

\section{Comparison to other models}
Deep CNNs have now becoming ubiquitous with image recognition and feature extraction, achieving state of the art in many major computer vision tasks. Therefore, like most other recent work, we also stick with CNNs for our visual model. However instead of training the the network from scratch, we use a publicly available model \textbf{OverFeat} which achieved very good performance in ILSVRC-2013. Our initial reason for doing this was to not waste extra time training the model from scratch on large data sets like ImageNet. They train the model with various special tricks like multiple-image fragments, multiple-scales, maximizing image localization, introducing color and translation invariance with data augmentation. Because of this the model provides excellent feature vector representation of images, and we feel that our good results on our task are partly because of the high modeling strength and accuracy of our visual model.

As explained above, we try several approaches with the textual model: simple BoW model, $n$-gram model, and SSE model. However our approach is much simpler than many researchers who have recently started using recurrent or recursive neural networks to model sentences \cite{model-SDT_RNN-socher,model-brnn-feifei}. Also, unlike \cite{model-defrag-feifei} we don't divide our images and sentences into fragments, but treat them as whole. With our results we show that these complicated models, which are also harder to train and slower, are not necessary to achieve good performance on the task of image-sentence bidirectional retrieval. 

Furthermore, we show an important feature of these bi-directional retrieval models: that they can be specialized to one task (either image search or image retrieval). We use empirical results to show that the methodology of generating negative sample (hereby referred to as \textit{training methodology}) has a significant on the final performance of the model (see Section \ref{sec:training_methodology}). If we use the I2T approach the model specializes in image annotation, and if we use T2I it specializes in image search. Note that it still remains bi-directional and gives good performance for the other task as well. 
\cleardoublepage
\chapter{Experiments and Results}\label{ch:experiments}

\section {Preprocessing}

The images were all scaled to a constant dimension of $221 \times 221$, since the ANN needs to have a constant sized input. This is also the input dimension of the OverFeat CNN which we utilize in our model. We do not crop image to make sure that a visually significant object is not cropped unintentionally, which will make it harder to identify it. 

We convert all sentences into lower-case, then remove punctuation and any other non-alphanumeric character from them, and tokenize them into a sequence of individual words. We filter out the articles "a", "an", and "the" since we believe they will not contribute to the semantic value of the sentence. Finally we limit the word vocabulary by number of occurrences. For unigram BoW, our vocabulary contains 5,000 words, while for bi-grams and tri-grams we use 50,000 vocabulary size. We experiment with treating these n-gram as both binary features and tf-idf (term frequency - inverse document frequency) features. The equation for calculating tf-idf is below:

$$\mbox{tf-idf}(w,d) = \frac{count(w,d)}{|w|} \times \log(\frac{|D|}{count(w,D)})$$

Here, $count(w,d)$ is the occurrence frequency of word $w$ in a document $d$, while $count(w,D)$ is the number of documents containing the word $w$. $|d|$ is the size of the vocabulary, while $|D|$ is the total number of documents. Tf-idf values have the advantage of storing relevance information, for example if a word  is frequent in all documents it gets a low tf-idf score. This can be used to filter out stop words.

\section{Evaluation Metrics}\label{evaluation_metrics}

In this sections below we will report the results of our models on sets of mainly Flickr30K data set and occasionally FLickr8K .Recall@K (\textbf{R@K}) is the most widely used metric in this domain.It is defined as the percentage of test queries for which a model returns the positive item among the top $k$ results. It is useful in the context of search where a user may be satisfied with the first $k$ results containing a single relevant item.

However, this definition is ambiguous if there are more than one positive results for each test query. This is true for $image annotation$ in our case, since each image corresponds with 5 sentences. Some researchers only consider the first out of 5 sentences in test cases (e.g. \cite{model-seminal_paper-hodosh, model-week_annotation-hodosh}) while others include all 5 (e.g. \cite{model-defrag-feifei}). We will take both into account, and report the following evaluation metrics:
\begin{itemize}
\item \textbf{R@K $\boldsymbol{1^{st}\_txt}$:} This is the Recall@K taking into consideration only the first out of five sentences for each image
\item \textbf{R@K $\boldsymbol{rnd\_txt}$:} This is the Recall@K taking into consideration one random sentence (determined at test time) for each image.
\item \textbf{R@K $\boldsymbol{avg\_txt}$:} This is the Recall@K taking into consideration all five of the sentences for each image one by one (iteratively), and then taking average over them. This is the most robust metric, since it does not depend on the relative position of sentence, and takes all sentences into account. Hence we use it for comparing our models.
\item \textbf{R@K $\boldsymbol{any\_txt}$:} This is the Recall@K taking into consideration any one of the five sentences. It returns a match if the image matches any one of the five sentences. This is only valid for $image caption$  because it needs multiple positive results for each query.
\item \textbf{med $\boldsymbol{r}$:} It is the value of $k$ for which the $R@K$ is equal to 50\%. In search intuition, it is the number of results to be displayed for making the probability of correct result appearing in the search results exactly 50\%. When calculating med $r$ we use Recall@K: $avg\_txt$.
\item \textbf{rPrecision(5):} It is the percentage of relevant items among the top 5 responses returned by the system. We select 5 because that is the maximum number of positive responses for $image annotation$. Since image search has only one relevant result, this metric is not relevant for it.
\end{itemize}

\section{Training Details}

We train our model in an end-to-end fashion using stochastic gradient descent (SGD) without momentum and weight decay terms. In all our experiments we fix the parameters of OverFeat network so its weights don't update with back propagation. In general it takes 1-2 days (depending on the size of the model) to fully train our model on the Flickr30K data set, using a CPU with 8 cores (16 threads). Typically the network runs for 50 epochs before termination. 

Out of the Flickr8K and FLickr30K data set we separate random 1000 images (And corresponding sentences) as the test set, and 5\% of the remaining examples as a held-out validation set for model selection. We tried several models and selected the meta-parameters based on performance on the held-out validation set. We select the parameters mentioned in \ref{tab:meta_parameters} for all our experiments for fair comparison.

\begin{table}[!ht]
\centering
\begin{tabular}{|c|c|}
\hline
\textbf{Parameter}   & \textbf{Value}   \\ \hline
margin ($\Delta$)     & 0.15    \\ \hline
learning rate ($lr$)       & 0.001  \\ \hline
lr decay                & linear \tablefootnote{$lr$ decreases linearly to a factor of 0.01 over 100 epochs}  \\ \hline
embedding dim. $n_{emb}$           & 300 - 1000    \\ \hline
non-linearity       & $relu$ \tablefootnote{rectified linear units}  \\ \hline
\end{tabular}
\caption{Hyper parameters showing best performance on held-out validation set (1,539 image and 7,695 sentences).}
\label{tab:meta_parameters}
\end{table}

\section{Experimental Results}

\begin{figure}[!ht]
\centering
\includegraphics[width=0.7\textwidth]{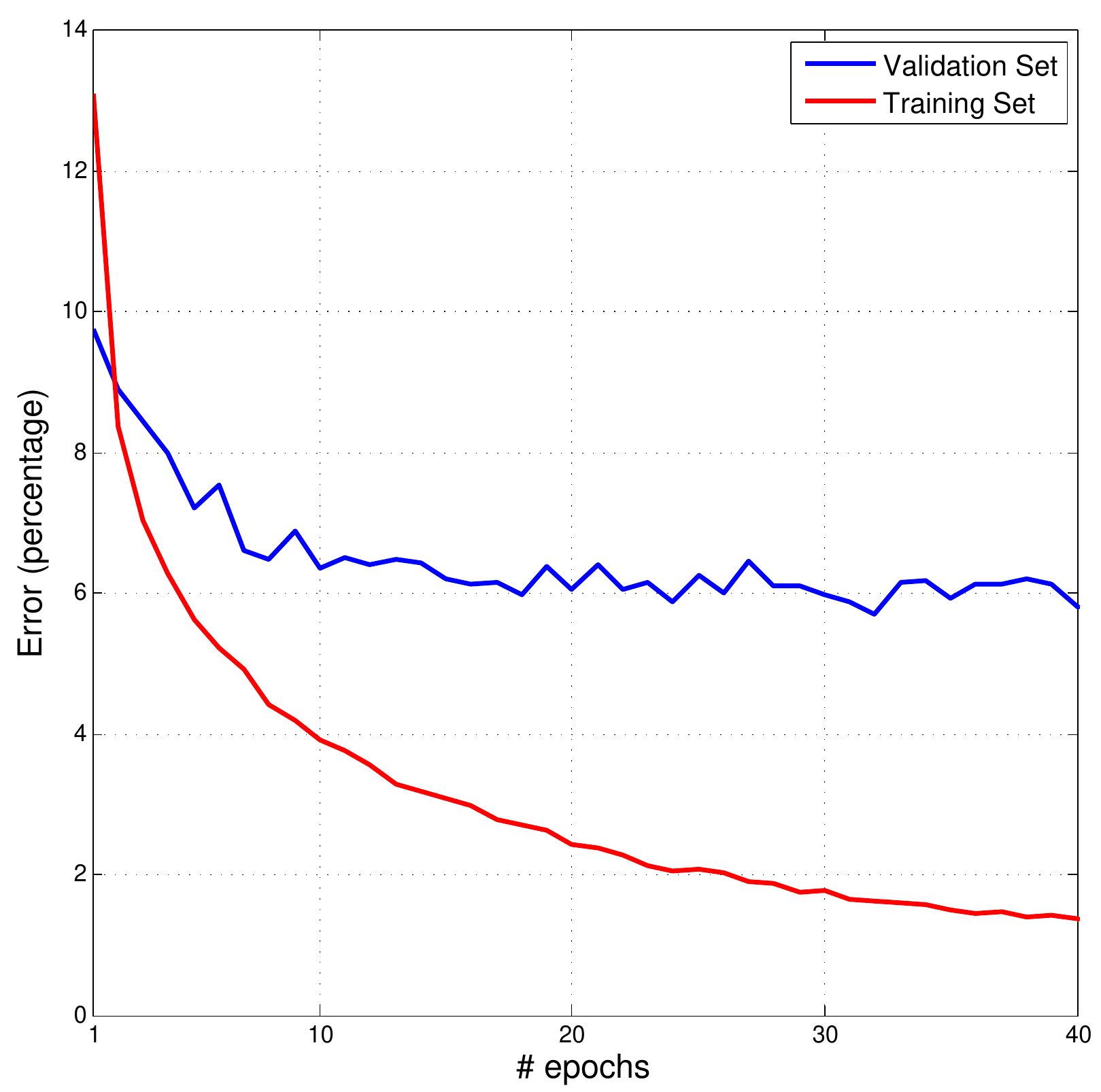}
\caption{Training time error on the \textbf{Flickr30K} data set. Textual model is a unigram BoW based ANN with 1 hidden layer and $n_{emb}$ of 1000. Error is the percentage of samples for which the model gives higher score to the positive sample over the negative sample. 50\% error specifies random performance.}
\label{fig:flickr30k_error_good}
\end{figure}

First of all, we would give a figure to show the error performance during training. Figure \ref{fig:flickr30k_error_good} shows that our model performs well on the validation set (with error going below 6\%). This is in comparison to our very first model which failed to generalize on the validation set, as seen in Figure \ref{fig:flickr8k_error_bad}. We can also see that the performance on validation set saturates around the $20^{th}$ epoch, and we only save the networks to the point that the performance does not saturate yet on the validation set. This is done to minimize the chance of overfitting on the training set.

\subsection{I2T and T2I training approach}

As explained in Chapter \ref{sec:training_methodology} we tried to methodologies for generating negative samples at training time, namely: \textbf{I2T} and \textbf{T2I}. We notice that this choice actually does impact the performance of our system in a peculiar way; it specializes it to a particular task. As is apparent from Table \ref{tab:Flickr30K_comparison_T2IvsI2T}, using I2T methodology gives higher performance for image annotation tasks and T2I methodology gives higher performance for image annotation task. We compare the performance across several models architectures and consistently notice the same phenomenon. Hence for all future results, whenever we refer to \textbf{image annotation} we report \textbf{I2T} results, and for \textbf{image search} we report \textbf{T2I} results.

Intuitively, the reason for this specialization is that I2T approach more closely matches with image annotation and T2I with image search. In I2T the images in both positive and negative samples are the same, while the sentences are different, so the system learns to be better at recognizing changes in sentences and hence ultimately performs better at image annotation. Vice versa holds for T2I and image search. It is, however, important to note that even after this specialization the each individually trained system is bi-directional and can do both tasks with reasonable accuracy (as seen from Table \ref{sec:training_methodology}.

To all look at more results (and other evaluation metrics for comparison) for comparing I2T and T2I methodology, please look at Table \ref{tab:T2IvsI2T_imageAnnotation} and Table \ref{tab:T2IvsI2T_imageSearch} in Appendix.

\begin{table}[!ht]
\centering
\begin{tabular}{|c|c|c|c|c|}
\hline
    \multicolumn{5}{|c|}{\textbf{Training methodology: I2T vs T2I}} \\  \hline
Models   & \multicolumn{2}{|c|}{image annotation}     & \multicolumn{2}{|c|}{image search} \\ \hline
  & I2T   & T2I    & I2T    & T2I  \\  \hline
$n_{hu}=300, n_{emb}=1000$, BoW   & \textbf{41.3}     & 39.34     & 37.92     & \textbf{41.08}   \\ \hline
$n_{hu}=300, n_{emb}=1000$, BoW w/o word2vec   & \textbf{40.48}     & 38.94    & 36.92    & \textbf{40.34}   \\ \hline
$n_{hu}=0, n_{emb}=300$, 2g   & 41.36     & 41.28     & 40.2     & \textbf{42.06}   \\ \hline
$n_{hu}=0, n_{emb}=1000$, 2g   & 43.34     & 42.84     & 40.2     & \textbf{43.74}   \\ \hline
$n_{hu}=1000, n_{emb}=1000$, 2g   & \textbf{45.4}     & 44     & 40.54     & \textbf{42.66}   \\ \hline
\end{tabular}
\caption{Comparison between two different training methodologies: I2T and T2I. \textit{BoW} means unsing unigram word features and \textit{2g} means using bigram. All results are given in Recall@10: $avg\_txt$. $n_{hu}$ is the hidden layer in the textual model, while $n_{emb}$ is the size of embedding dimension.}
\label{tab:Flickr30K_comparison_T2IvsI2T}
\end{table}

\subsection{Results of BoW model}

First we will report the results of our BoW model, in which our textual model is simply an MLP with unigram word features. All information about temporal dependency in the sentence is lost and we use a dictionary size of 5,000 most frequently words (after removing "a", "an" and "the"). We also use binary features so there is no information about multiple occurrences of words in a sentence.

The hidden layer of textual model is fixed to 300 units, since the word2vec embeddings have dimensionality of 300. We fixed the hidden layer of the trainable visual model to 1,000 units, and the size of multimodal embedding vector is also fixed to 1,000. Table \ref{tab:Flickr8K} shows the results on Flickr8K test set, and Table \ref{tab:Flickr30K_unigram} shows results on Flickr30K test set (both of size 1,000 images). We use this model named as \textbf{our model: BoW} when comparing results in Chapter \ref{ch:results}.

Inspecting the tables, the results look quite promising, and we are ready to explore more architectures and parameters.

\begin{table}[!ht]
\centering
\begin{tabular}{c|c|c|c|c|c|c|c|c}
\hline
\multicolumn{9}{|c|}{\textbf{BoW model on Flickr8K}} \\  \hhline{|=|========|}
\multicolumn{1}{|c|}{\textbf{Metric} } & \multicolumn{4}{|c|}{\textbf{Image Annotation}}    & \multicolumn{4}{|c|}{\textbf{Image Search}}    \\ \hhline{|=|====|====|}
\multicolumn{1}{|c|}{rPrecision(5)}   & \multicolumn{4}{|c|}{7.32}    & \multicolumn{4}{|c|}{-}    \\ \hline
\multicolumn{1}{|c|}{med $r$}     & \multicolumn{4}{|c|}{21}      & \multicolumn{4}{|c|}{22}  \\  \hhline{|=|====|====|}
\multicolumn{1}{|c|}{\textbf{k}}   & \textbf{1}     & \textbf{2}     & \textbf{5}     & \textbf{10}     & \textbf{1}     & \textbf{2}     & \textbf{5}     & \multicolumn{1}{|c|}{\textbf{10}} \\  \hhline{|=|=|=|=|=|=|=|=|=|}
\multicolumn{1}{|c|}{R@K: $rnd\_txt$}   & 8.7     & 12     & 22.5     & 33.7     & 6.1     & 14.7     & 25.9     & \multicolumn{1}{|c|}{37.4} \\ \hline
\multicolumn{1}{|c|}{R@K: $avg\_txt$}   & 7.38     & 12.5     & 23.72     & 35.22     & 7.3     & 13.44     & 25.5     & \multicolumn{1}{|c|}{37.74} \\ \hline
\end{tabular}
\caption{Flickr8K experiments with unigram BoW model and Word2Vec initialization. \textbf{R@K} is Recall@K (high is good). \textbf{Med $\boldsymbol{r}$} is the median rank (low is good). \textbf{rPrecision(5)} is the precision at rank 5 (high is good).}
\label{tab:Flickr8K}
\end{table}

\begin{table}[!ht]
\centering
\begin{tabular}{c|c|c|c|c|c|c|c|c}
\hline
\multicolumn{9}{|c|}{\textbf{our model: BoW}} \\  \hhline{|=|========|}
\multicolumn{1}{|c|}{\textbf{Metric} } & \multicolumn{4}{|c|}{\textbf{Image Annotation}}    & \multicolumn{4}{|c|}{\textbf{Image Search}}    \\ \hhline{|=|====|====|}
\multicolumn{1}{|c|}{rPrecision(5)}   & \multicolumn{4}{|c|}{10.2}    & \multicolumn{4}{|c|}{-}    \\ \hline
\multicolumn{1}{|c|}{med $r$}     & \multicolumn{4}{|c|}{17}      & \multicolumn{4}{|c|}{16}  \\  \hhline{|=|====|====|}
\multicolumn{1}{|c|}{\textbf{k}}   & \textbf{1}     & \textbf{2}     & \textbf{5}     & \textbf{10}     & \textbf{1}     & \textbf{2}     & \textbf{5}     & \multicolumn{1}{|c|}{\textbf{10}} \\  \hhline{|=|=|=|=|=|=|=|=|=|}
\multicolumn{1}{|c|}{R@K: $rnd\_txt$}   & 10.6     & 14.9     & 29.2     & 42.7     & 9.8     & 19.3     & 28.9     & \multicolumn{1}{|c|}{40.5} \\ \hline
\multicolumn{1}{|c|}{R@K: $avg\_txt$}   & 10.4     & 16.94     & 29.3     & 41.3     & 10.56     & 18.04     & 29.56     & \multicolumn{1}{|c|}{41.08} \\ \hline
\multicolumn{1}{|c|}{R@K: $1^{st}\_txt$}   & 10.7     & 17.5     & 30.4     & 43.8     & 9.6     & 18.2     & 30.6     & \multicolumn{1}{|c|}{44.2} \\ \hline
\multicolumn{1}{|c|}{R@K: $any\_txt$}   & 14.0     & 19.0     & 33.0     & 45.5     & -     & -    & -     & \multicolumn{1}{|c|}{-} \\ \hline
\end{tabular}
\caption{Flickr30K experiments with unigram BoW model and Word2Vec initialization. \textbf{R@K} is Recall@K (high is good). \textbf{Med $\boldsymbol{r}$} is the median rank (low is good). \textbf{rPrecision(5)} is the precision at rank 5 (high is good).}
\label{tab:Flickr30K_unigram}
\end{table}

\subsection{Word2Vec results}

Next we investigate the effect of Word2Vec embeddings. As stated earlier, we used these as a way to have good initialization of our textual model. Since we don't train our model on a large corpus of text, we speculate word2vec would be helpful. We repeat the experiment on Flickr30K using BoW model, with the only difference that we use randomly initialized network. Doing this experiment is useful because we suggested using n-gram models as a second step onto BoW model, however we are aware of no such publicly available n-gram embeddings.

The results of the experiment are shown in Table \ref{tab:Flickr30K_noWord2Vec} which we can compare with Table \ref{tab:Flickr30K_unigram}. A quick comparison between the performance metrics in these two tables shows that although word2vec initialization does help in improving the accuracy of the system, but the effects are not extremely drastic. For example R@10:$avg\_txt$ for image annotation increases by 3.3\% while for image search increases by 3.86\%. 

The results are significant in suggesting the importance of using word2vec for initializing our models, but we feel that the gap may be bridged with using more complicated models without this initialization.

\begin{table}[!ht]
\centering
\begin{tabular}{c|c|c|c|c|c|c|c|c}
\hline
\multicolumn{9}{|c|}{\textbf{BoW model w/o Word2Vec}} \\  \hhline{|=|========|}
\multicolumn{1}{|c|}{\textbf{Metric} } & \multicolumn{4}{|c|}{\textbf{Image Annotation}}    & \multicolumn{4}{|c|}{\textbf{Image Search}}    \\ \hhline{|=|====|====|}
\multicolumn{1}{|c|}{rPrecision(5)}   & \multicolumn{4}{|c|}{10.08}    & \multicolumn{4}{|c|}{-}    \\ \hline
\multicolumn{1}{|c|}{med $r$}     & \multicolumn{4}{|c|}{17}      & \multicolumn{4}{|c|}{18}  \\  \hhline{|=|====|====|}
\multicolumn{1}{|c|}{\textbf{k}}   & \textbf{1}     & \textbf{2}     & \textbf{5}     & \textbf{10}     & \textbf{1}     & \textbf{2}     & \textbf{5}     & \multicolumn{1}{|c|}{\textbf{10}} \\  \hhline{|=|=|=|=|=|=|=|=|=|}
\multicolumn{1}{|c|}{R@K: $rnd\_txt$}   & 11.2     & 16.6     & 31.4     & 41.2     & 10.5     & 17.6     & 28.4     & \multicolumn{1}{|c|}{39.8} \\ \hline
\multicolumn{1}{|c|}{R@K: $avg\_txt$}   & 10.36     & 17.08     & 29.5     & 40.48     & 10.12     & 17.26     & 29.04     & \multicolumn{1}{|c|}{40.34} \\ \hline
\end{tabular}
\caption{Flickr30K experiments without Word2Vec initialization. \textbf{R@K} is Recall@K (high is good). \textbf{Med $\boldsymbol{r}$} is the median rank (low is good). \textbf{rPrecision(5)} is the precision at rank 5 (high is good).}
\label{tab:Flickr30K_noWord2Vec}
\end{table}

\subsection{Results of n-gram models}

We replaced the BoW textual model with n-gram model. N-gram models conserve some sentence-order information and so theoretically they act as better features. In our experiments we tried bigrams (\textbf{2g}), trigrams (\textbf{3g}) and trigrams + skip grams (\textbf{tk3}). In all these cases the vocabulary size is fixed to 50,000 which is the size of input layer of textual model MLP. Table \ref{tab:Flickr30K_comparison_ngramType} shows a comparison between these three different kind of models for our task.

The results show that both bigrams and trigrams perform better in different settings, while inclusion of skip grams in trigram model consistently resulted in worst performance.

\begin{table}[!ht]
\centering
\begin{tabular}{|c|c|c|c|c|c|c|}
\hline
    \multicolumn{7}{|c|}{\textbf{n-gram model comparison}} \\  \hline
Models   & \multicolumn{6}{|c|}{Evaluation Metric}      \\ \hline
    & \multicolumn{3}{|c|}{\textbf{rPrecision(5)}}      & \multicolumn{3}{|c|}{\textbf{R@10: $\boldsymbol{avg\_txt}$}}  \\  \hline
  & \textbf{2g}    & \textbf{3g}    & \textbf{tk3}    & \textbf{2g}    & \textbf{3g}    & \textbf{tk3} \\  \hline
$n_{hu}=300, n_{emb}=1000$  & 10.12     & \textbf{11.08}     & 9.5     & 40.72     & \textbf{42.7}     & 38.46  \\ \hline
$n_{hu}=1000, n_{emb}=300$   & 10.52     & \textbf{11.34}     & 9.78    & 41.46     & 41.54     & 38.82      \\ \hline
$n_{hu}=1000, n_{emb}=1000$   & 11.68     & 11.0     & 10.7     & \textbf{43.36}     & 41.74     & 40.04     \\ \hline
$n_{hu}=0, n_{emb}=1000$   & \textbf{11.96}    & 11.04     & 10.6     & \textbf{43.34}     & 40.96     & 41.62   \\ \hline
$n_{hu}=0, n_{emb}=300$   & \textbf{12.2}    & 11.9     & 10.56    & 41.36    & \textbf{42.24}    & 41.16    \\ \hline
\end{tabular}
\caption{Comparison between different $n$-gram models, on Flickr30K data. \textbf{2g} is bi-gram, \textbf{3g} is trigram and \textbf{tk3} is combination of tri-gram and skip-gram features. In all case size of input vocabulary is 50,000 and all rest model parameters are similar. The bold figures show best results in the specific row. $n_{hu}$ is the hidden layer in the textual model, while $n_{emb}$ is the size of embedding dimension.}
\label{tab:Flickr30K_comparison_ngramType}
\end{table}

\bigskip
Following this we also experimented on using either binary or tf-idf valued features. Intuition suggests that tf-idf encode more information and so should perform better than binary features. However, empirical evidence suggests otherwise as binary features performed better in most of our models. Hence for future models we stick with binary features. The results are shown in Table \ref{tab:Flickr30K_comparison_wordFeatures}

\begin{table}[!ht]
\centering
\begin{tabular}{|c|c|c|c|c|}
\hline
    \multicolumn{5}{|c|}{\textbf{n-gram features comparison}} \\  \hline
Models   & \multicolumn{4}{|c|}{Evaluation Metric}      \\ \hline
    & \multicolumn{2}{|c|}{\textbf{rPrecision(5)}}      & \multicolumn{2}{|c|}{\textbf{R@K: $\boldsymbol{avg\_txt}$}}  \\  \hline
  & \textbf{binary}    & \textbf{tf-idf}    & \textbf{binary}    & \textbf{tf-idf}  \\  \hline
$n_{hu}=0, n_{emb}=300$, 2g   & \textbf{12.2}     & 11.58     & 41.36     & 42.2   \\ \hline
$n_{hu}=0, n_{emb}=300$, 3g   & 11.9     & 11.7    & \textbf{42.24}    & 41.84   \\ \hline
$n_{hu}=0, n_{emb}=300$, tk3   & 10.56     & 10.3     & \textbf{41.16}    & 39.42    \\ \hline
$n_{hu}=1000, n_{emb}=300$, 3g   & \textbf{11.34}    & 10.28     & 41.54     & \textbf{41.76}   \\ \hline
$n_{hu}=1000, n_{emb}=1000$, 2g   & \textbf{11.68}    & 10.52     & 43.36    & 41.86  \\ \hline
$n_{hu}=1000, n_{emb}=1000$, 3g  & 11.0     & 11.04     & \textbf{41.74}     & 40.96   \\ \hline
$n_{hu}=1000, n_{emb}=1000$, tk3 & 10.7     & 10.58     & 40.04     & 40.1   \\ \hline
\end{tabular}
\caption{Comparison between different input word features: $binary$ and $tf-idf$, on Flickr30K data. In all case size of input vocabulary is 50,000 and the inputs are either 2g, 3g, or tk3, and all rest model parameters are similar. The bold figures show best results in the specific row. $n_{hu}$ is the hidden layer in the textual model, while $n_{emb}$ is the size of mbedding dimension.}
\label{tab:Flickr30K_comparison_wordFeatures}
\end{table}

\bigskip
Finally we select the best performing model as having 1,000 unit hidden layers in both textual and trainable visual models, and also 1,000 unit multimodal embedding vector. The results of this model on Flickr30K are shown in Figure \ref{tab:Flickr30K_bigram_2layer}, and we notice significant improvements over the BoW model. We use this model named as \textbf{our model: n-gram} when comparing results in Chapter \ref{ch:results}.

\begin{table}[!ht]
\centering
\begin{tabular}{c|c|c|c|c|c|c|c|c}
\hline
    \multicolumn{9}{|c|}{\textbf{our model: n-gram}} \\  \hhline{|=|========|}
\multicolumn{1}{|c|}{\textbf{Metric} } & \multicolumn{4}{|c|}{\textbf{Image Annotation}}    & \multicolumn{4}{|c|}{\textbf{Image Search}}    \\ \hhline{|=|====|====|}
\multicolumn{1}{|c|}{rPrecision(5)}   & \multicolumn{4}{|c|}{11.68}    & \multicolumn{4}{|c|}{-}    \\ \hline
\multicolumn{1}{|c|}{med $r$}     & \multicolumn{4}{|c|}{15}      & \multicolumn{4}{|c|}{15}  \\  \hhline{|=|====|====|}
\multicolumn{1}{|c|}{\textbf{k}}   & \textbf{1}     & \textbf{2}     & \textbf{5}     & \textbf{10}     & \textbf{1}     & \textbf{2}     & \textbf{5}     & \multicolumn{1}{|c|}{\textbf{10}} \\  \hhline{|=|=|=|=|=|=|=|=|=|}
\multicolumn{1}{|c|}{R@K: $rnd\_txt$}   & 12.3     & 19.2     & 32     & 44.4     & 11.7     & 20.4     & 29.3     & \multicolumn{1}{|c|}{44.4} \\ \hline
\multicolumn{1}{|c|}{R@K: $avg\_txt$}   & 11.62     & 18.88     & 31.44     & 43.36     & 11.44     & 18.56     & 31.24     & \multicolumn{1}{|c|}{42.66} \\ \hline
\multicolumn{1}{|c|}{R@K: $1^{st}\_txt$}   & 13.0     & 19.7     & 32.6     & 45.4     & 11.9     & 20.9     & 34.9     & \multicolumn{1}{|c|}{44.3} \\ \hline
\multicolumn{1}{|c|}{R@K: $any\_txt$}   & 14.9     & 24.7     & 39.3     & 50.9     & -      & -    & -     & \multicolumn{1}{|c|}{-} \\ \hline
\end{tabular}
\caption{Flickr30K experiments with bigram model and binary inputs. \textbf{R@K} is Recall@K (high is good). \textbf{Med $\boldsymbol{r}$} is the median rank (low is good). \textbf{rPrecision(5)} is the precision at rank 5 (high is good).}
\label{tab:Flickr30K_bigram_2layer}
\end{table}

\section{Are deep textual models necessary}\label{deep_nets_necessary}

Now we will, investigate if a deep network is really necessary in the textual model. We explored several shallow single layered architectures, where the input n-gram input is linearly transformed into the embedding vector. Surprisingly, almost all of the networks we tried reached performance close to the ones we achieved with a deeper network with non-linearities. In addition to performing well, these shallow models were approximately $3\times$ faster to train (ignoring the time required for OverFeat feature vector generation - since it is not fine tuned, it is equivalent to only single time pass)

We reached the best performance using a 300 dimensional embedding (linearly connected with 5,000 input units). The small trainable visual network also does not contain any hidden layer as well, and the 4,096 dimensional OverFeat feature vectors are linearly transformed into the 300 embedding dimension. The result of this model is in Table \ref{tab:Flickr30K_bigram_1layer}. Comparison can be made with Table \ref{tab:Flickr30K_bigram_2layer} to see the performance difference as compared to the earlier discussed \textit{our model: n-gram}. We use this model named as \textbf{our model: shallow} when comparing results in Chapter \ref{ch:results}.

\begin{table}[!ht]
\centering
\begin{tabular}{c|c|c|c|c|c|c|c|c}
\hline
    \multicolumn{9}{|c|}{\textbf{our model: shallow}} \\  \hhline{|=|========|}
\multicolumn{1}{|c|}{\textbf{Metric} } & \multicolumn{4}{|c|}{\textbf{Image Annotation}}    & \multicolumn{4}{|c|}{\textbf{Image Search}}    \\ \hhline{|=|====|====|}
\multicolumn{1}{|c|}{rPrecision(5)}   & \multicolumn{4}{|c|}{12.2}    & \multicolumn{4}{|c|}{-}    \\ \hline
\multicolumn{1}{|c|}{med $r$}     & \multicolumn{4}{|c|}{17}      & \multicolumn{4}{|c|}{16}  \\  \hhline{|=|====|====|}
\multicolumn{1}{|c|}{\textbf{k}}   & \textbf{1}     & \textbf{2}     & \textbf{5}     & \textbf{10}     & \textbf{1}     & \textbf{2}     & \textbf{5}     & \multicolumn{1}{|c|}{\textbf{10}} \\  \hhline{|=|=|=|=|=|=|=|=|=|}
\multicolumn{1}{|c|}{R@K: $rnd\_txt$}   & 11.2     & 19.5     & 30.4     & 42.3     & 12.7     & 19.6     & 30.2     & \multicolumn{1}{|c|}{42.2} \\ \hline
\multicolumn{1}{|c|}{R@K: $avg\_txt$}   & 12.16     & 19.36     & 30.94     & 42.06     & 12.16     & 19.36     & 30.94     & \multicolumn{1}{|c|}{42.06} \\ \hline
\multicolumn{1}{|c|}{R@K: $1^{st}\_txt$}   & 13.7     & 21.7     & 32.9     & 42.9     & 13.2     & 20.5     & 33.7     & \multicolumn{1}{|c|}{44.0} \\ \hline
\multicolumn{1}{|c|}{R@K: $any\_txt$}   & 16.0     & 24.1     & 39.5     & 50.7     & -     & -    & -      & \multicolumn{1}{|c|}{-} \\ \hline
\end{tabular}
\caption{Flickr30K experiments with bigram model and binary inputs. \textbf{R@K} is Recall@K (high is good). \textbf{Med $\boldsymbol{r}$} is the median rank (low is good). \textbf{rPrecision(5)} is the precision at rank 5 (high is good).}
\label{tab:Flickr30K_bigram_1layer}
\end{table}

\bigskip
Lastly, we also explored using the SSE (semantic sequence embedding) module as our textual model. Details of architecture are given in Chapter \ref{ch:proposed_model}. It has specialized architecture for modeling sequence inside variable length sentences, and similar architectures have achieved good performance in task such as language modeling \cite{text-unified-collobert} and review based sentiment analysis\cite{text-SSE}. We were hopeful that it would result in performance boost on our task as well.

However, empirical results show another story and the model performs not as well as other previously discussed models. The results of the best model architecture are shown in Table \ref{tab:Flickr30K_sse}. The architecture has context window of size 5 and both word level and phrase level embeddings of 300 dimensions. This model is initialized with word2vec, and it is important to note that if we did not use word2vec initialization the performance fell staggeringly to about half the current performance.

\bigskip
We feel a possible reason of the failure of SSE as our textual module is because there isn't enough textual training data. Our data set is limited to 150,000 sentences (5 sentences for each of approximately 30,000 images). This makes a total of approximately 1.5-2 million words. In contrast \cite{text-unified-collobert} use 631 million words to train their language model with similar architecture. The SSE architecture is much deeper and has more parameters than other textual models we tried, it has 4 distinct layers: lookup table with individual word embeddings (which we initialized with word2vec), CNN with fixed window for phrase level embeddings, averaging layer for complete sentence layer embedding and finally a non-linearity followed by a linear layer for greater modeling capacity. Because of this deep architecture, it is not unlikely that it needs more multimodal data to train well. 

Secondly, the fact that it word2vec boosts its performance (validation error at train time saturates at about 15\% if not using word2vec and at about 8\% if using word2vec in most models) also means that it is unable to learn good word embeddings if initialized from random. In contrast when we initialized our BoW model without word2vec, the difference in performance was not much slighter. This again reinforces our opinion that more data is needed.

\begin{table}[!ht]
\centering
\begin{tabular}{c|c|c|c|c|c|c|c|c}
\hline
    \multicolumn{9}{|c|}{\textbf{SSE for textual model}} \\  \hhline{|=|========|}
\multicolumn{1}{|c|}{\textbf{Metric} } & \multicolumn{4}{|c|}{\textbf{Image Annotation}}    & \multicolumn{4}{|c|}{\textbf{Image Search}}    \\ \hhline{|=|====|====|}
\multicolumn{1}{|c|}{rPrecision(5)}   & \multicolumn{4}{|c|}{7.84}    & \multicolumn{4}{|c|}{-}    \\ \hline
\multicolumn{1}{|c|}{med $r$}     & \multicolumn{4}{|c|}{21}      & \multicolumn{4}{|c|}{24}  \\  \hhline{|=|====|====|}
\multicolumn{1}{|c|}{\textbf{k}}   & \textbf{1}     & \textbf{2}     & \textbf{5}     & \textbf{10}     & \textbf{1}     & \textbf{2}     & \textbf{5}     & \multicolumn{1}{|c|}{\textbf{10}} \\  \hhline{|=|=|=|=|=|=|=|=|=|}
\multicolumn{1}{|c|}{R@K: $rnd\_txt$}   & 6.6     & 15.2     & 23.9     & 37.1     & 7.4     & 14.3     & 22.8     & \multicolumn{1}{|c|}{32.5} \\ \hline
\multicolumn{1}{|c|}{R@K: $avg\_txt$}   & 7.58     & 13.18     & 24.54     & 35.82     & 7.98     & 13.62     & 24.16     & \multicolumn{1}{|c|}{34.5} \\ \hline
\multicolumn{1}{|c|}{R@K: $1^{st}\_txt$}   & 8.0     & 13.5     & 24.6     & 35.3     & 7.7     & 13.4     & 23.8     & \multicolumn{1}{|c|}{35.8} \\ \hline
\multicolumn{1}{|c|}{R@K: $any\_txt$}   & 9.6     & 16.1     & 27.2     & 39.8     & -     & -    & -      & \multicolumn{1}{|c|}{-} \\ \hline
\end{tabular}
\caption{Flickr30K experiments with SSE textual model (window size =5). \textbf{R@K} is Recall@K (high is good). \textbf{Med $\boldsymbol{r}$} is the median rank (low is good). \textbf{rPrecision(5)} is the precision at rank 5 (high is good).}
\label{tab:Flickr30K_sse}
\end{table}

\section{Comparison with existing systems}

We would like to compare the performance of our model with existing state of the art systems. However, we noted in Chapter \ref{evaluation_metrics} that there is inherent ambiguity in the way the most common evaluation metric (Recall@K) is defined for this task. In addition to this, researchers often differ how they present the results.  We use all 5 sentences while evaluating our system. For comparing \textbf{image annotation} task we will report \textbf{R@K $\boldsymbol{avg\_txt}$} and for comparing \textbf{image search} task we will report \textbf{R@K $\boldsymbol{any\_txt}$}. Here we will follow the standard by \cite{model-defrag-feifei}, and this is how they report it to the best of our knowledge.

\begin{table}[!ht]
\centering
\begin{tabular}{|c|c|c|c|c|c|c|}
\hline
    \multicolumn{7}{|c|}{\textbf{Flickr30K: Comparison}} \\  \hline
Models   & \multicolumn{3}{|c|}{image annotation}     & \multicolumn{3}{|c|}{image search} \\ \hline
  &\textbf{R@1}   &\textbf{R@5}    &\textbf{R@10}    &\textbf{R@1}    &\textbf{R@5}   &\textbf{R@10}  \\ \hhline{|=|======|}
Random Ranking  & 0.1     & 0.5     & 1.0     & 0.1     & 0.5       & 1.0   \\ \hhline{|=|======|}
DeVise \tablefootnote{\cite{model-devise-samy}}   & 4.5     & 18.1    & 29.2    & 6.7   &21.9   &32.7   \\ \hline
SDT-RNN \tablefootnote{\cite{model-SDT_RNN-socher}}  &9.6     &29.8     &41.1    &8.9  &29.8  &41.1   \\ \hline
DeFrag \tablefootnote{\cite{model-defrag-feifei}} FAO\tablefootnote{fragment alignment objective} &11.0   &28.7     &39.3     &7.6  &23.8  &34.5   \\ \hline
DeFrag GO\tablefootnote{global ranking objective}   &11.5   &33.2     &44.9     &8.8  &27.6  &38.4   \\ \hline
DeFrag FAO + GO     &12.0   &37.1     &50.0     &9.9  &30.5  &43.2   \\ \hline
(*)DeFrag FAO + GO + MIL\tablefootnote{multi instance learning}    &14.2   &37.7     &51.3     &10.2  &30.8  &44.2   \\ \hline
DeFrag (*) + Finetune CNN    &16.4   &40.2  &54.7     &10.3  &31.4  &44.5   \\ \hhline{|=|======|}
our model: BoW  &14.0   &33.0   &45.5  &10.56  &29.56  &41.08 \\ \hline
our model: n-gram &14.9  &39.3  &50.9  &11.44  &31.24  &42.66 \\ \hline
our model: shallow &16.0  &39.5  &50.7 & 12.16     & 30.94     & 42.06 \\ \hline  
\end{tabular}
\caption{Comparison between existing image text matching systems. \textbf{R@K} is Recall@K (high is good). In our models, for image annotation we report R@K: $any\_txt$ and for image search we report R@K: $avg\_txt$.}
\label{tab:final comparison}
\end{table}

\bigskip
Table \ref{tab:final comparison} shows that our models are comparable to recent work in the field \footnote{Some of these models have been re-implemented by \cite{model-defrag-feifei}}, and perform not far from the top results. When comparing the results it is interesting to note that most recent research has gone into complicated modeling on the textual domain. For example \cite{model-SDT_RNN-socher} use SDT-RNN which is dependency-tree based recursive neural network. They also try other types of recurrent and recursive neural networks. Besides this \cite{model-defrag-feifei} use dependency tree edges to fragment  their sentences, and so on. In contrast we use simple textual model and achieve comparable results. The bulk of our model's strength comes from the visual model and the training methodology which specializes the performance towards a particular task boosting its performance. In our experiments we notice that temporal dependency (or information about word order) modeled with simple bi-grams or tri-grams is enough to achieve good results.
\cleardoublepage
\chapter{Conclusion}\label{ch:results}
The goal of the project was to explore deep learning architectures in the context of multimodal image and text modeling. Our task was based on bi-directional retrieval. This means that our model would support retrieving images based on a text query, and also retrieving text sentences based on an image query. Applications of such a system would include automatic image annotation, and sentence based image retrieval. To design our system we use the ranking approach which is common to many traditional information retrieval systems. At each iteration we feed our image with a positive pair and a negative pair of training input. A positive input is a matched pair of image and sentence (as labeled by a human), while a negative input is simply a mismatched pair. The model assigns a score to the positive input and to the negative input, and the objective function is to make the positive sample score higher than the negative sample. In this way the model will learn to rank related pair of image and text higher than an arbitrary unrelated pair.

\bigskip
Since images and text have very different statistical properties, it is necessary to model two different sub-models for them. The idea is that if both sub-models extract underlying semantic information (such as information about the content) and encode it into a distributed vector embedding, then a ranking objective function could align these two embeddings into a common multimodal space. Once this is the case, we can simply compute the distance betwen the two vectors as a measure of similarity. For the visual model we explored using deep convolutional neural network (CNN). Our initial model was initialized from random weights and it failed to perform well on our tasks. We did some experiments on different but related task of image recognition on MNIST and Caltech256, and were able to conclude lack of pretraining as a reason why the network failed. Following this we adopted OverFeat into our visual model, which is a publicly available CNN which achieved high performance on ILSVRC 2013. With just this change the network stats to work well on the task, which confirms our initial suspicion. For our textual module we tried various models and features - some simple and some more complicated. We explored using bag of words approach as well as n-grams (bigrams, trigrams and skip grams), and tried binary and tf-idf features. We experimented with using shallow linear architectures, intermediate architecture like MLP and deeper architectures with word embeddings and convolutional neural network (for modeling variable length sentence by averaging over a fixed window). 

\bigskip
We evaluate the performance of our models by testing them on Flickr8K and Flickr30K training sets. A comparison with recent work in the field shows that even though our models are simpler, we achieve performance comparable to some of the top results. We believe that a major reason for our good performance is the performance of the CNN, which extracts very powerful and semantically relevant features. Although like other approaches, OverFeat is trained on the ImageNet data base, but it is trained with special tricks like multiple-image fragments, multiple-scales, maximizing image localization, introducing color and translation invariance with data augmentation. Because of this the network achieves excellent performance on ILSVRC 2013 \cite{resource-overfeat} challenge, and this also increases the quality of feature vectors extracted by the model.

\bigskip
For the textual module we observed that even a very shallow module with a single linear layer achieves very good performance. Although using hidden layers and increasing the dimensionality of the embedding layer does enhance the performance, but performance is still good with a simple linear layer. This provides an interesting contrast between the visual module and the textual module. Since only a very well pretrained CNN was able to train on the task. However a more specialized and deeper architecture (SSE) does not perform so well on the textual side. This provides a fundamental question of how essential is deep learning in this task? We think that the only reason CNN performs well is that it was pretrained with a lot of data to extract semantic features from images. So in order to make a textual model work well we either need a very large scale image-text data base, or we need to pretrain the deeper textual model on tasks which allow it to extract semantic information from sentences. An example of such a task could be document classification.

\bigskip
Finally, we show an important feature of our bi-directional retrieval models: that they can be specialized to one task (either image search or image retrieval). We use empirical results to show that the methodology of generating negative sample has a significant on the final performance of the model. If we use the I2T approach the model specializes in image annotation, and if we use T2I it specializes in image search. Note that it still remains bi-directional and gives good performance for the other task as well.

\section{Future work}

The simplest addition to our model would be to turn on updating for our CNN while we are training on the image-text data set. This is called finetuning, since the CNN is pretrained on image data alone. We expect to achieve performance improvement by this finetuning, as reported by \cite{model-defrag-feifei} and seen in the Table \ref{tab:final comparison}. However this will come at a cost of significantly additional training time, hence it is a trade off.

\bigskip
As noted in Chapter \ref{deep_nets_necessary}, we experimented with a more promising model SSE but experienced a performance drop to our surprise. We hypothesize that this is because the complicated and deeper model needs a larger training data to perform well, something which is far from true with our Flickr30K and Flickr8K data sets. Interestingly, this problem is similar to what we faced initially with our visual model and explained in Chapter \ref{ch:methodology}. Our uninitialized CNN failed at the tasks miserably, which convinced us to use a CNN pretrained on ImageNet. We similarly feel that an SSE module trained on large relevant data would significantly improve our results. We experienced a portion of this when - as we reported - initializing the first layer of the SSE with word2vec caused the accuracy of the model to almost double (albeit still lesser than our simpler models). This means that although word2vec are good word embeddings, but they are not enough for the task of multimodal learning since they were initially learned on just a unimodal language model. The best approach would be to use a large image-text data base to train our models, however a high quality large scale data base does not exist in this task yet. Regeradless, there are some data sets that we aim to use in the future.

\bigskip
MS-COCO\footnote{http://mscoco.org/} is a data set which provides high quality captions labeled by 5 different humans, for more than 100,000 images. Although still not an extremely large data set, it is the first we hope to try our models on. Some other larger data sets exist but their usability in image text mapping systems has been questioned. Flickr1M \footnote{http://press.liacs.nl/mirflickr/} is an example of such data set which contains 1 million images from Flickr website associated with captions added by people uploading the images. \cite{model-seminal_paper-hodosh} argue that such captions do not act in good association with images, since humans tend to describe the aspects of the images which are not so apparent from the image itself. In contrast, our system only wants to know what is apparent inside the image.
\cite{model-week_annotation-hodosh} explored using a good quality dataset (Flickr30K) to improve performance on a low quality annotated data sets (SBU1M and Flickr1M). We propose that it would be interesting to explore the opposite: to pretrain our model with large amount of weakly annotated text and then finetuning it on data sets like Flickr30K and MS-COCO. 

\bigskip
Another exciting area to explore is using data augmentation similar to how it is being used in visual models for image recognition. In other words we could artificially augment the textual data (number of sentences) matching each image in our training set. There can be three ways of doing this. Firstly we could use the original human labeled sentences as seeds and use text based information retrieval systems to retrieve similar sentences from a large textual corpus. Secondly we could design a language model to generate more sentences similar to the one it is given as input. Thirdly, we could use an image-based sentence generation system to generate additional relevant sentences and use these when training our retrieval system. Once we have a larger amount of text corresponding to images we can better train deeper and high-capacity models, and also better model inherent variance in describing an image.

\bigskip
The next step is to expand the system with some ideas presented above. These include, finetuning our CNN, pretraining entire system on a weekly annotated large scale data set, and artificially augmenting textual data. This is our first experimentation with image text multimodal modeling system, and we aim to keep on improving it.

\cleardoublepage
\begin{appendices}

\FloatBarrier
\chapter{Acronyms}
\begin{acronym}
\acro{ANN}{artificial neural network}
\acro{BoW}{bag of words}
\acro{CNN}{convolutional neural network}
\acro{I2T}{image to text}
\acro{KCCA}{kernel conanical correlation analysis}
\acro{R@K}{recall at k}
\acro{MLP}{multi layer perceptron}
\acro{NLP}{natural language Procecesing}
\acro{relu}{rectified linear unit}
\acro{RNN}{recurrent neural network}
\acro{RNN}{recursive neural network}
\acro{SGD}{stochastic gradient descent}
\acro{SSE}{supervised sequence embedding}
\acro{SSI}{supervised semantic indexing}
\acro{T2I}{text to image}
\acro{tf-tdf}{term frequency - inverse document frequency}
\end{acronym}

\FloatBarrier
\chapter{Extra Resources}

\section{MNIST}
\textbf{MNIST} \footnote{http://yann.lecun.com/exdb/mnist/} is a database of hand-written digits consisting of 60,000 training examples and 10,000 test examples. The digits are size-normalized and centered in a fixed-size image of $28\times28$ pixels. The data set has a long history of acting as a benchmark for measuring performance of visual recognition systems.

\section{Caltech256}
\textbf{Caltech256} \footnote{http://www.vision.caltech.edu/Image\_Datasets/Caltech256/} is a database for object recognition from images. It consists of about 30,000 images belonging to 256 categories (plus clutter). Categories range from car, helicopter, airplane, dog, cat, elephant, spaghetti, rifle, and other everyday objects.  The images are not left-right aligned which makes the task relatively harder. 

\section{ImageNet}
\textbf{ImageNet} \footnote{http://www.image-net.org/} is a large scale image database organized in a hierarchical manner.  It consists of a total of about 14 million images belonging to 22,000 categories. In addition to this they organize a yearly competition for computer vision called ILSVRC (Image Net Large Scale Visual Recognition Challenge). The competition consists of three individual tasks: object recognition, localization, and detection.

\section{Pascal VOC 2008}
This was the first publicly available data set purposefully collected for the image text ranking task by \cite{resource-PascalVOC} This data set consists of 1,000 images from PASCAL VOC-2008 object recognition challenge. 50 images are randomly selected belonging to each of the 20 categories. Each image is annotated with 5 descriptive captions using Amazon's Mechanical Turk, resulting in 5,000 sentences.

Although, being the first specific data set, it suffers from a number of shortcomings which limit its usefulness. The domain of its images is very limited and  the captions are relatively simple and sometimes containing grammatical and spelling errors \cite{model-seminal_paper-hodosh}.

\FloatBarrier
\chapter{Software}
We trained the neural network models using proprietary machine leaning environment developed by NEC Labs America. MiLDe (Machine Learning Development Environment) is a software environment for developing and prototyping applications based on Machine Learning and statistics.

\FloatBarrier
\chapter{Further results of I2T and T2I training methodologies}

\begin{table}[!ht]
\centering
\begin{tabular}{c|c|c|c|c|c|c|c|c}
\hline
\multicolumn{9}{|c|}{\textbf{Image Annotation}} \\  \hhline{|=|========|}
\multicolumn{1}{|c|}{\textbf{Metric} } & \multicolumn{4}{|c|}{\textbf{I2T training apprch.}}    & \multicolumn{4}{|c|}{\textbf{T2I training apprch.}}    \\ \hhline{|=|====|====|}
\multicolumn{1}{|c|}{rPrecision(5)}   & \multicolumn{4}{|c|}{10.2}    & \multicolumn{4}{|c|}{9.66}    \\ \hline
\multicolumn{1}{|c|}{med $r$}     & \multicolumn{4}{|c|}{17}      & \multicolumn{4}{|c|}{19}  \\  \hhline{|=|====|====|}
\multicolumn{1}{|c|}{\textbf{k}}   & \textbf{1}     & \textbf{2}     & \textbf{3}     & \textbf{4}     & \textbf{1}     & \textbf{2}     & \textbf{3}     & \multicolumn{1}{|c|}{\textbf{4}} \\  \hhline{|=|=|=|=|=|=|=|=|=|}
\multicolumn{1}{|c|}{R@K: $rnd\_txt$}   & 10.6     & 14.9     & 29.2     & 42.7     & 8.2     & 14.9     & 27.1     & \multicolumn{1}{|c|}{35.9} \\ \hline
\multicolumn{1}{|c|}{R@K: $avg\_txt$}   & 10.4     & 16.94     & 29.3     & 41.3     & 9.48     & 15.76     & 26.8     & \multicolumn{1}{|c|}{37.92} \\ \hline
\multicolumn{1}{|c|}{R@K: $1^{st}\_txt$}   & 10.7     & 17.5     & 30.4     & 43.8     & 10.8     & 16.4     & 27.2     & \multicolumn{1}{|c|}{39.7} \\ \hline
\multicolumn{1}{|c|}{R@K: $any\_txt$}   & 14.0     & 19.0     & 33.0     & 45.5     & 12.6     & 18.9    & 31.9     & \multicolumn{1}{|c|}{42.3} \\ \hline
\end{tabular}
\caption{Experiment shows training approach significantly effects results, int this case \textbf{I2T approach} boosts performance for \textbf{image annotation}. Experiment performed with Flickr30K data set with unigram BoW model and Word2Vec initialization. rPrecision(5) high is good,  Med $r$ low is good, and R@K high is good.}
\label{tab:T2IvsI2T_imageAnnotation}
\end{table}

\begin{table}[!ht]
\centering
\begin{tabular}{c|c|c|c|c|c|c|c|c}
\hline
\multicolumn{9}{|c|}{\textbf{Image Search}} \\  \hhline{|=|========|}
\multicolumn{1}{|c|}{\textbf{Metric} } & \multicolumn{4}{|c|}{\textbf{T2I training apprch.}}    & \multicolumn{4}{|c|}{\textbf{I2T training apprch.}}    \\ \hhline{|=|====|====|}
\multicolumn{1}{|c|}{med $r$}     & \multicolumn{4}{|c|}{16}      & \multicolumn{4}{|c|}{18}  \\  \hhline{|=|====|====|}
\multicolumn{1}{|c|}{\textbf{k}}   & \textbf{1}     & \textbf{2}     & \textbf{3}     & \textbf{4}     & \textbf{1}     & \textbf{2}     & \textbf{3}     & \multicolumn{1}{|c|}{\textbf{4}} \\  \hhline{|=|=|=|=|=|=|=|=|=|}
\multicolumn{1}{|c|}{R@K: $rnd\_txt$}   & 9.8     & 19.3     & 28.9     & 40.5     & 10.1     & 15.8     & 27.3     & \multicolumn{1}{|c|}{39.3} \\ \hline
\multicolumn{1}{|c|}{R@K: $avg\_txt$}   & 10.56     & 18.04     & 29.56     & 41.3     & 9.48     & 15.6     & 27.9     & \multicolumn{1}{|c|}{39.34} \\ \hline
\multicolumn{1}{|c|}{R@K: $1^{st}\_txt$}   & 9.6     & 18.2     & 30.6     & 44.2     & 8.7     & 16.1     & 30.3     & \multicolumn{1}{|c|}{42.6} \\ \hline
\end{tabular}
\caption{Experiment shows training approach significantly effects results, int this case \textbf{T2I approach} boosts performance for \textbf{image search}. Experiment performed with Flickr30K data set with unigram BoW model and Word2Vec initialization. Med $r$ low is good, and R@K high is good.}
\label{tab:T2IvsI2T_imageSearch}
\end{table}
	
\end{appendices}

\bibliographystyle{apalike}
\addcontentsline{toc}{chapter}{Bibliography}
\bibliography{./library}

\begin{thebibliography}{}

\bibitem[Bach and Jordan, 2003]{bach2003kernel}
Bach, F.~R. and Jordan, M.~I. (2003).
\newblock Kernel independent component analysis.
\newblock {\em The Journal of Machine Learning Research}, 3:1--48.

\bibitem[Bai et~al., 2009]{text-SSI}
Bai, B., Weston, J., Grangier, D., Collobert, R., Sadamasa, K., Qi, Y.,
  Chapelle, O., and Weinberger, K. (2009).
\newblock Supervised semantic indexing.
\newblock In {\em Proceedings of the 18th ACM conference on Information and
  knowledge management}, pages 187--196. ACM.

\bibitem[Bengio et~al., 2003]{text-neural_language_model-yoshua}
Bengio, Y., Ducharme, R., Vincent, P., and Janvin, C. (2003).
\newblock A neural probabilistic language model.
\newblock {\em The Journal of Machine Learning Research}, 3:1137--1155.

\bibitem[Bespalov et~al., 2011]{text-SSE}
Bespalov, D., Bai, B., Qi, Y., and Shokoufandeh, A. (2011).
\newblock Sentiment classification based on supervised latent n-gram analysis.
\newblock In {\em Proceedings of the 20th ACM international conference on
  Information and knowledge management}, pages 375--382. ACM.

\bibitem[Collobert and Weston, 2008]{text-unified-collobert}
Collobert, R. and Weston, J. (2008).
\newblock A unified architecture for natural language processing: Deep neural
  networks with multitask learning.
\newblock In {\em Proceedings of the 25th international conference on Machine
  learning}, pages 160--167. ACM.

\bibitem[Deng et~al., 2009]{resources-imagenet-feifei}
Deng, J., Dong, W., Socher, R., Li, L.-J., Li, K., and Fei-Fei, L. (2009).
\newblock Imagenet: A large-scale hierarchical image database.
\newblock In {\em Computer Vision and Pattern Recognition, 2009. CVPR 2009.
  IEEE Conference on}, pages 248--255. IEEE.

\bibitem[Farhadi et~al., 2010]{resource-PascalVOC}
Farhadi, A., Hejrati, M., Sadeghi, M.~A., Young, P., Rashtchian, C.,
  Hockenmaier, J., and Forsyth, D. (2010).
\newblock Every picture tells a story: Generating sentences from images.
\newblock In {\em Computer Vision--ECCV 2010}, pages 15--29. Springer.

\bibitem[Frome et~al., 2013]{model-devise-samy}
Frome, A., Corrado, G.~S., Shlens, J., Bengio, S., Dean, J., Mikolov, T.,
  et~al. (2013).
\newblock Devise: A deep visual-semantic embedding model.
\newblock In {\em Advances in Neural Information Processing Systems}, pages
  2121--2129.

\bibitem[Gong et~al., 2014]{model-week_annotation-hodosh}
Gong, Y., Wang, L., Hodosh, M., Hockenmaier, J., and Lazebnik, S. (2014).
\newblock Improving image-sentence embeddings using large weakly annotated
  photo collections.
\newblock In {\em Computer Vision--ECCV 2014}, pages 529--545. Springer.

\bibitem[Gupta et~al., 2012]{gupta2012choosing}
Gupta, A., Verma, Y., and Jawahar, C. (2012).
\newblock Choosing linguistics over vision to describe images.
\newblock In {\em AAAI}.

\bibitem[Hinton et~al., 2006]{hinton2006fast}
Hinton, G., Osindero, S., and Teh, Y.-W. (2006).
\newblock A fast learning algorithm for deep belief nets.
\newblock {\em Neural computation}, 18(7):1527--1554.

\bibitem[Hinton et~al., 2012]{hinton2012dropout}
Hinton, G.~E., Srivastava, N., Krizhevsky, A., Sutskever, I., and
  Salakhutdinov, R.~R. (2012).
\newblock Improving neural networks by preventing co-adaptation of feature
  detectors.
\newblock {\em arXiv preprint arXiv:1207.0580}.

\bibitem[Hodosh et~al., 2013]{model-seminal_paper-hodosh}
Hodosh, M., Young, P., and Hockenmaier, J. (2013).
\newblock Framing image description as a ranking task: Data, models and
  evaluation metrics.
\newblock {\em Journal of Artificial Intelligence Research}, pages 853--899.

\bibitem[Hubel and Wiesel, 1968]{hubel1968receptive}
Hubel, D.~H. and Wiesel, T.~N. (1968).
\newblock Receptive fields and functional architecture of monkey striate
  cortex.
\newblock {\em The Journal of physiology}, 195(1):215--243.

\bibitem[Karpathy and Fei-Fei, 2014]{model-brnn-feifei}
Karpathy, A. and Fei-Fei, L. (2014).
\newblock Deep visual-semantic alignments for generating image descriptions.
\newblock {\em arXiv preprint arXiv:1412.2306}.

\bibitem[Karpathy et~al., 2014]{model-defrag-feifei}
Karpathy, A., Joulin, A., and Li, F. F.~F. (2014).
\newblock Deep fragment embeddings for bidirectional image sentence mapping.
\newblock In {\em Advances in Neural Information Processing Systems}, pages
  1889--1897.

\bibitem[Kiros et~al., 2014]{model-multimodalNLM-ruslan}
Kiros, R., Salakhutdinov, R., and Zemel, R. (2014).
\newblock Multimodal neural language models.
\newblock In {\em Proceedings of the 31st International Conference on Machine
  Learning (ICML-14)}, pages 595--603.

\bibitem[Krizhevsky et~al., 2012]{vision-krizhevsky}
Krizhevsky, A., Sutskever, I., and Hinton, G.~E. (2012).
\newblock Imagenet classification with deep convolutional neural networks.
\newblock In {\em Advances in neural information processing systems}, pages
  1097--1105.

\bibitem[Le~Cun et~al., 1990]{le1990handwritten}
Le~Cun, B.~B., Denker, J.~S., Henderson, D., Howard, R.~E., Hubbard, W., and
  Jackel, L.~D. (1990).
\newblock Handwritten digit recognition with a back-propagation network.
\newblock In {\em Advances in neural information processing systems}. Citeseer.

\bibitem[LeCun et~al., 1998]{vision-document-lenet5-lecun}
LeCun, Y., Bottou, L., Bengio, Y., and Haffner, P. (1998).
\newblock Gradient-based learning applied to document recognition.
\newblock {\em Proceedings of the IEEE}, 86(11):2278--2324.

\bibitem[LeCun et~al., 1995]{vision-comparison-lecun}
LeCun, Y., Jackel, L., Bottou, L., Brunot, A., Cortes, C., Denker, J., Drucker,
  H., Guyon, I., Muller, U., Sackinger, E., et~al. (1995).
\newblock Comparison of learning algorithms for handwritten digit recognition.
\newblock In {\em International conference on artificial neural networks},
  volume~60, pages 53--60.

\bibitem[Lin, 2004]{lin2004rouge}
Lin, C.-Y. (2004).
\newblock Rouge: A package for automatic evaluation of summaries.
\newblock In {\em Text Summarization Branches Out: Proceedings of the ACL-04
  Workshop}, pages 74--81.

\bibitem[Lin et~al., 2014]{lin2014microsoft}
Lin, T.-Y., Maire, M., Belongie, S., Hays, J., Perona, P., Ramanan, D.,
  Doll{\'a}r, P., and Zitnick, C.~L. (2014).
\newblock Microsoft coco: Common objects in context.
\newblock In {\em Computer Vision--ECCV 2014}, pages 740--755. Springer.

\bibitem[Lowe, 2004]{lowe2004distinctive}
Lowe, D.~G. (2004).
\newblock Distinctive image features from scale-invariant keypoints.
\newblock {\em International journal of computer vision}, 60(2):91--110.

\bibitem[Makadia et~al., 2010]{makadia2010baselines}
Makadia, A., Pavlovic, V., and Kumar, S. (2010).
\newblock Baselines for image annotation.
\newblock {\em International Journal of Computer Vision}, 90(1):88--105.

\bibitem[Miikkulainen and Dyer, 1991]{miikkulainen1991natural}
Miikkulainen, R. and Dyer, M.~G. (1991).
\newblock Natural language processing with modular pdp networks and distributed
  lexicon.
\newblock {\em Cognitive Science}, 15(3):343--399.

\bibitem[Mikolov et~al., 2013a]{resource-word2vec-related}
Mikolov, T., Chen, K., Corrado, G., and Dean, J. (2013a).
\newblock Efficient estimation of word representations in vector space.
\newblock {\em CoRR}, abs/1301.3781.

\bibitem[Mikolov et~al., 2013b]{resource-word2vec}
Mikolov, T., Sutskever, I., Chen, K., Corrado, G., and Dean, J. (2013b).
\newblock Distributed representations of words and phrases and their
  compositionality.
\newblock {\em CoRR}, abs/1310.4546.

\bibitem[Ordonez et~al., 2011]{ordonez2011im2text}
Ordonez, V., Kulkarni, G., and Berg, T.~L. (2011).
\newblock Im2text: Describing images using 1 million captioned photographs.
\newblock In {\em Advances in Neural Information Processing Systems}, pages
  1143--1151.

\bibitem[Papineni et~al., 2002]{papineni2002bleu}
Papineni, K., Roukos, S., Ward, T., and Zhu, W.-J. (2002).
\newblock Bleu: a method for automatic evaluation of machine translation.
\newblock In {\em Proceedings of the 40th annual meeting on association for
  computational linguistics}, pages 311--318. Association for Computational
  Linguistics.

\bibitem[Rashtchian et~al., 2010]{resource-Flickr8K}
Rashtchian, C., Young, P., Hodosh, M., and Hockenmaier, J. (2010).
\newblock Collecting image annotations using amazon's mechanical turk.
\newblock In {\em Proceedings of the NAACL HLT 2010 Workshop on Creating Speech
  and Language Data with Amazon's Mechanical Turk}, pages 139--147. Association
  for Computational Linguistics.

\bibitem[Reiter and Belz, 2009]{reiter2009investigation}
Reiter, E. and Belz, A. (2009).
\newblock An investigation into the validity of some metrics for automatically
  evaluating natural language generation systems.
\newblock {\em Computational Linguistics}, 35(4):529--558.

\bibitem[Sch{\"o}lkopf et~al., ]{scholkopfgreedy}
Sch{\"o}lkopf, B., Platt, J., and Hofmann, T.
\newblock Greedy layer-wise training of deep networks.

\bibitem[Sermanet et~al., 2014]{resource-overfeat}
Sermanet, P., Eigen, D., Zhang, X., Mathieu, M., Fergus, R., and LeCun, Y.
  (2014).
\newblock Overfeat: Integrated recognition, localization and detection using
  convolutional networks.
\newblock In {\em International Conference on Learning Representations (ICLR
  2014)}. CBLS.

\bibitem[Socher et~al., 2014]{model-SDT_RNN-socher}
Socher, R., Karpathy, A., Le, Q.~V., Manning, C.~D., and Ng, A.~Y. (2014).
\newblock Grounded compositional semantics for finding and describing images
  with sentences.
\newblock {\em Transactions of the Association for Computational Linguistics},
  2:207--218.

\bibitem[Socher et~al., 2011]{text-RNN-socher}
Socher, R., Lin, C.~C., Manning, C., and Ng, A.~Y. (2011).
\newblock Parsing natural scenes and natural language with recursive neural
  networks.
\newblock In {\em Proceedings of the 28th international conference on machine
  learning (ICML-11)}, pages 129--136.

\bibitem[Srivastava and Salakhutdinov, 2012]{model-multimodalDBM-ruslan}
Srivastava, N. and Salakhutdinov, R.~R. (2012).
\newblock Multimodal learning with deep boltzmann machines.
\newblock In {\em Advances in neural information processing systems}, pages
  2222--2230.

\bibitem[Vinyals et~al., 2014]{model-NIC-samy}
Vinyals, O., Toshev, A., Bengio, S., and Erhan, D. (2014).
\newblock Show and tell: A neural image caption generator.
\newblock {\em arXiv preprint arXiv:1411.4555}.

\bibitem[Young et~al., 2014]{resource-Flickr30K}
Young, P., Lai, A., Hodosh, M., and Hockenmaier, J. (2014).
\newblock From image descriptions to visual denotations: New similarity metrics
  for semantic inference over event descriptions.
\newblock {\em Transactions of the Association for Computational Linguistics},
  2:67--78.

\end{thebibliography}

\end{document}